%% file: 00main.tex
\documentclass[11pt]{article}

\usepackage{times}
\usepackage{latexsym}

\usepackage[T1]{fontenc}

\usepackage[utf8]{inputenc}

\usepackage{microtype}

\usepackage{inconsolata}

\usepackage{graphicx}

\usepackage{amsmath}
\usepackage{booktabs}
\usepackage{multirow}
\usepackage{placeins}
\usepackage{subcaption}
\usepackage{siunitx}

\usepackage[final]{acl}

\title{Evaluating LLM-as-a-Judge Beyond Score Alignment: A Psychometric Analysis of Residual Judging Difficulty}

\author{
  \textbf{Longwei Cong\textsuperscript{1}},
  \textbf{Sonja Hahn\textsuperscript{1}},
  \textbf{Sebastian Gombert\textsuperscript{1}},
  \textbf{Leon Camus\textsuperscript{1}}, \\
  \textbf{Fabian Zehner\textsuperscript{1,2}},
  \textbf{Hendrik Drachsler\textsuperscript{1,3}},
  \textbf{Ulf Kroehne\textsuperscript{1,4}}
  \\
  \textsuperscript{1}DIPF | Leibniz Institute for Research and Information in Education
  \\
  \textsuperscript{2}Centre for International Student Assessment (ZIB)
  \\
  \textsuperscript{3}Faculty of Computer Science, Goethe University Frankfurt
  \\
  \textsuperscript{4}Chemnitz University of Technology
  \\
  \small{\texttt{\{l.cong,s.hahn,s.gombert,l.camus,f.zehner,h.drachsler,u.kroehne\}@dipf.de}}
}

\begin{document}
\maketitle

\begin{abstract}
Large language models (LLMs) are widely used as automatic judges, with validity typically assessed via alignment with human scores. However, aggregate agreement fails to reveal whether humans and LLMs find the same evaluation cases difficult. In this paper, we study this problem in summarization evaluation from a psychometric perspective. We fit Many-Facet Rasch Models separately to human and LLM ratings to decompose scores into latent summary quality, rater severity, dimension severity, and rating-scale thresholds. Building on this decomposition, we define residual hardness as a model-adjusted measure of judging difficulty and compare whether human and LLM judges share the same hardness structure.

Across 17 open-weight LLM judges on SummEval, we find that moderate alignment in latent summary quality does not imply alignment in residual hardness. Human and LLM judges differ in which summary--dimension units remain difficult, and this mismatch is strongly dimension-dependent. Consistency shows a pronounced LLM-hard shift, whereas coherence shows a human-hard shift. We further show that human-easy but LLM-hard cases are partially predictable from observable source--summary properties. These findings suggest that aggregate human alignment reflects only part of LLM-as-a-judge reliability, while psychometric residual diagnostics support more informative judge evaluation and more targeted human--LLM collaboration.

\end{abstract}

\input{01introduction}
\input{02background}
\input{03method}
\input{04results}
\input{05discusstion}
\input{06conclution}
\input{07limitation}

\FloatBarrier

\bibliography{custom.bib}

\appendix
\input{08appendix}

\end{document}

%% file: 01introduction.tex
\section{Introduction}

Large language models (LLMs) have attracted substantial attention for their strong performance across a wide range of natural language processing tasks \cite{naveed2025comprehensive}. Their advanced language understanding and reasoning capabilities have also enabled their use as automatic judges for complex tasks, offering a scalable alternative or complement to traditional expert-driven evaluation \cite{gu2025surveyllmasajudge}. LLM-as-a-judge methods have been applied to social intelligence evaluation \cite{wang-etal-2024-sotopia}, multimodal evaluation \cite{chen2024mllm}, and domain-specific settings such as finance \cite{yu2024fincon}, law \cite{cheong2024not}, and education \cite{Cong2026ASAG}. In this setting, an LLM is prompted to assign scores, rankings, labels, or pairwise preferences according to task-specific criteria or natural-language rubrics. The appeal is that LLMs could approximate aspects of human judgment while overcoming the cost, time, and scalability bottlenecks inherent in human evaluation \cite{zheng2023judging}. Consequently, prior work typically validates LLM judges by measuring their alignment with human judgments, most often through correlations with human ratings or agreement with human preference labels \cite{gu2025surveyllmasajudge}.

However, agreement on average quality provides only a coarse view of judging behavior and reveals little about the difficulty structure of evaluation~\citep{rodriguez-etal-2021-evaluation,cong2026estimatingllmgradingability}. Two judge groups may assign similar overall scores while struggling with different evaluation instances. This distinction is important because an automatic judge should not only reproduce aggregate human scores on benchmarks, but also behave reliably on individual cases in real-world evaluation settings \cite{li-etal-2025-generation,zhou2025lost}. If humans and LLMs disagree about which cases are difficult to judge, aggregate alignment metrics may overstate the reliability of LLM-as-a-judge systems.

We argue that a shift toward a more measurement-oriented perspective is essential for a deeper understanding of LLM judging behavior. Rather than evaluating only whether LLM judges reproduce human scores on average, we need diagnostics that distinguish alignment in output quality from alignment in judging difficulty.

To address this gap, we draw on Item Response Theory (IRT), a psychometric framework for modeling observed ratings in terms of latent traits and systematic measurement facets. Specifically, we use a rating-scale Many-Facet Rasch Model (MFRM) \cite{eckes2023introduction} because it separates latent item quality from systematic rating effects such as rater severity, dimension-specific severity, and rating-scale thresholds. 
Building on this model, we define residual hardness as the average absolute residual between observed ratings and the ratings expected by the fitted MFRM. Conceptually, this metric provides a post-fit diagnostic of local judging instability after accounting for the systematic facets of the evaluation process. It therefore distinguishes low estimated summary quality from large residual deviations in the rating process, without treating residual hardness as an intrinsic latent measure of item difficulty.

We study this diagnostic in a SummEval case study of summarization evaluation, a setting in which each system summary is rated along multiple dimensions, such as coherence, consistency, fluency, and relevance, that place different demands on judges. This makes summarization a useful testbed for analyzing whether human--LLM alignment extends beyond aggregate quality scores to the structure of judging difficulty \cite{fabbri-etal-2021-summeval,liu2023geval}.

Our analysis is guided by three research questions.

\paragraph{RQ1}
Does alignment between human raters and LLM judges in latent summary quality also imply alignment in residual hardness?

\paragraph{RQ2}
How does residual-hardness mismatch vary across evaluation dimensions?

\paragraph{RQ3}
Can textual features predict which summaries are easy for humans but difficult for LLMs to evaluate?

These questions are practically important for human--LLM collaboration scenarios. Cases that are likely to be human-easy but LLM-hard can be routed to human review, additional source-grounding checks, or stronger judge models~\citep{bondi2022role,10.1145/3613904.3641960}. A difficulty-aware view therefore supports more targeted human--LLM collaboration in evaluation.

The contributions of this paper are as follows:

\begin{itemize}
\item We propose a psychometrically grounded diagnostic that separates latent instance quality from residual judging difficulty, providing a more fine-grained view of LLM-as-a-judge behavior.
\item We show that human--LLM alignment in latent instance quality does not imply alignment in residual hardness, revealing substantial differences in the residual-hardness patterns of human and LLM judge panels.
\item We demonstrate that human-easy but LLM-hard cases are partially predictable from observable summary-, source-, and system-level features, supporting difficulty-aware human--LLM evaluation pipelines.
\end{itemize}

%% file: 02background.tex
\section{Background}

\subsection{LLM-as-a-judge and Human Alignment}

As complex evaluation tasks become increasingly difficult to assess with fixed references or surface-form metrics, recent work has turned to large language models as automatic judges \cite{gu2025surveyllmasajudge}. LLM judges have been used for general-purpose evaluation of model outputs, including summarization, dialogue, and instruction following \cite{liu2023geval,fu2024gptscore,zheng2023judging}, as well as for data annotation \cite{tan2024large} and domain-specific educational assessment \cite{gombert2026rubrics}.

The validity of LLM-as-a-judge systems is typically assessed by measuring alignment with human judgments. Prior work reports correlations with human quality ratings \cite{liu2023geval,fu2024gptscore}, rank correlations between systems \cite{liu2023geval,fu2024gptscore}, agreement with pairwise human preferences \cite{zheng2023judging,li2023alpacaeval,wang2024pandalm}, or agreement with aggregated human labels \cite{bavaresco2025llms}. These studies show that LLMs can serve as useful evaluators in some settings, but also that their reliability varies substantially across datasets, evaluated properties, judge models, and forms of human annotation \cite{bavaresco2025llms}.

This literature establishes human alignment as a central criterion for evaluating LLM judges. However, most existing alignment metrics focus on whether LLMs reproduce human scores, rankings, preferences, or labels. They provide less information about whether human and LLM judges share the same structure of judging difficulty. A judge may correlate with humans in aggregate quality assessment while still diverging on which evaluation instances are unstable, ambiguous, or difficult to rate. This distinction is important for model training \cite{yu-etal-2025-improve}, benchmark construction \cite{zhou2025lost}, and human--AI collaboration \cite{pan-etal-2024-human}, where the reliability of individual judgments matters as much as aggregate agreement. We therefore adopt a psychometric view of LLM-as-a-judge alignment that separates latent quality and systematic rating facets from post-fit local residual variation.

\subsection{Item Response Theory}

Item Response Theory (IRT) provides a measurement framework for modeling observed responses in terms of latent respondent characteristics and item properties \cite{van2016handbook}. The Rasch model is a foundational IRT model in which the probability of a correct response depends on the difference between a respondent's latent ability and an item's difficulty:
\[
P(y_{mn}=1 \mid \theta_m, b_n) = \sigma(\theta_m - b_n),
\]
where \(y_{mn}\in\{0,1\}\) denotes the observed response of respondent \(m\) to item \(n\), \(\theta_m\) denotes the latent ability of respondent \(m\), \(b_n\) denotes the difficulty of item \(n\), and \(\sigma(\cdot)\) is the logistic sigmoid.

The Many-Facet Rasch Model (MFRM) \cite{eckes2023introduction} extends the Rasch model by incorporating additional facets that may systematically influence observed scores. It allows the measurement model to include further sources of variation, such as rater severity, task difficulty, scoring dimensions, or category thresholds. 

Recent work has begun to apply IRT to LLM-as-a-judge evaluation in several ways \cite{ye2026largelanguagemodelpsychometrics}. Choi et al. use a graded-response IRT model to diagnose LLM-as-a-judge reliability, formalizing reliability in terms of prompt-level intrinsic consistency and alignment with human quality assessments \cite{choi2026diagnosing}. Recent educational assessment work has also applied many-facet Rasch measurement to evaluate whether LLM raters exhibit acceptable severity, consistency, and rater effects in automated scoring \citep{jiao2025comparinghumanairater, WANG2025100481}. These studies establish psychometric models as useful tools for assessing LLM raters. However, they primarily ask whether LLM raters are reliable and human-aligned at the score or rater level. In contrast, we analyze whether human annotators and LLM judges exhibit similar post-fit residual-hardness patterns. 

%% file: 03method.tex
\section{Method}

We first obtain ratings from a panel of LLM judges on the summaries and dimensions rated by human annotators. We then analyze these ratings with MFRM, modeling summary quality, rater severity, dimension effects, and rating-scale thresholds. Based on the fitted model, we compare human and LLM judges in latent measurement space and introduce a residual-based measure of judging hardness.

\subsection{Dataset and LLMs}

We use SummEval \cite{fabbri-etal-2021-summeval}, a multi-rater benchmark for evaluating generated summaries. SummEval was constructed in the context of news summarization evaluation and contains summaries generated for source articles from the CNN/DailyMail dataset. The human-annotated subset used in our study consists of 1,600 system summaries, obtained from 100 source documents and 16 summarization systems. Each evaluated unit corresponds to a system summary for a source document. The dataset includes summaries produced by both extractive and abstractive summarization systems. Human ratings cover four evaluation dimensions: coherence, consistency, fluency, and relevance. The original annotations include three expert annotators and five crowd annotators, with anonymized rater IDs. The primary human MFRM uses all eight human raters, comprising the three expert annotators and five crowd annotators.

To construct LLM-as-a-judge annotations, we prompt 17 open-weight language models to rate the same summary--dimension units on the same 1--5 scale. The models span multiple developers and model families, including Google \cite{Kamath2025Gemma3T}, Qwen \cite{qwen3.5}, Microsoft \cite{abdin2024phi, abdin2024phi3technicalreporthighly}, Meta-Llama \cite{grattafiori2024llama}, MistralAI \cite{liu2026ministral}, and OpenChat \cite{ICLR2024_fc8781fb}, with parameter sizes ranging from 0.8B to 14B. We choose this panel to obtain a diverse, reproducible, and locally runnable set of LLM judges. Decoding is performed greedily with \texttt{do\_sample=False} and temperature set to 0. Under this setting, token sampling is disabled, so changing the sampling seed does not produce independent generations. Details of the model set and the prompt are provided in Appendix \ref{app:llm-judge-prompt}.

\subsection{Many-Facet Rasch Measurement}

We fit rating-scale MFRM models separately for human raters and LLM raters. For a rating \(y_{rid}\) assigned by rater \(r\) to summary \(i\) on dimension \(d\), the adjacent-category logit is defined as:
\begin{equation}
\begin{split}
\log
\frac{P(y_{rid}=c+1)}
     {P(y_{rid}=c)}
&=
\theta_i - \rho_r - \delta_d - \tau_c,\\
&\qquad c=1,\ldots,4.
\end{split}
\end{equation}

Here, \(\theta_i\) denotes the latent quality of summary \(i\), \(\rho_r\) denotes the severity of rater \(r\), \(\delta_d\) denotes the severity associated with evaluation dimension \(d\), and \(\tau_c\) denotes the threshold between adjacent rating categories \(c\) and \(c+1\). Higher \(\theta_i\) indicates higher estimated summary quality. Higher \(\rho_r\) indicates a stricter rater, higher \(\delta_d\) indicates a dimension on which high scores are harder to obtain after controlling for summary quality and rater severity, and higher \(\tau_c\) indicates a more difficult transition to the next higher rating category. Details of the estimation procedure are provided in Appendix~\ref{app:mfrm_estimation}.

We assess the adequacy of the MFRM estimation using standard psychometric diagnostics. First, we examine standardized residuals and report MAE and RMSE as descriptive summaries of rating-level prediction error. Second, we report item separation reliability, which indicates whether the calibrated measures distinguish summaries with sufficient precision relative to measurement error. We additionally verify that all five response categories are used by both human and LLM raters.

\subsection{Residual Hardness}

Classical agreement metrics measure whether two groups assign similar scores, but they do not directly capture whether the same items exhibit larger local deviations after accounting for the main structure of the rating process. We use the term residual hardness as a post-fit diagnostic of local judging instability. In the MFRM, expected ratings are determined by summary quality, rater severity, dimension severity, and rating-scale thresholds. If ratings for a summary--dimension unit remain far from their expected values after these facets have been accounted for, the unit exhibits high residual hardness.

For each observed rating, the fitted MFRM yields an expected rating:
\begin{equation}
\hat{y}_{rid} = E[y_{rid}\mid \hat{\theta}_i,\hat{\rho}_r,\hat{\delta}_d,\hat{\tau}].
\end{equation}

The response residual is:
\begin{equation}
e_{rid} = y_{rid} - \hat{y}_{rid}.
\end{equation}

For a summary-dimension unit \((i,d)\), residual hardness is defined as the mean absolute residual over raters in group \(g\):
\begin{equation}
H^{(g)}_{id}
= \frac{1}{|R_g|}
\sum_{r\in R_g} |e_{rid}|.
\end{equation}

Residual hardness is not low summary quality or a conventional latent item-difficulty parameter. Rather, it summarizes the magnitude of local deviations from the fitted MFRM after accounting for the modeled facets. We therefore interpret it as a post-fit diagnostic of local judging instability rather than as an intrinsic latent measure of item difficulty, ambiguity, or uncertainty. Larger values may reflect local rater disagreement, model misfit, or remaining unmodeled structure. The human and LLM analyses use the same MFRM form but separately estimated parameters, and we compare whether the resulting panels exhibit similar residual-hardness patterns. Appendix~\ref{app:hardness_sanity_check} relates residual hardness to local rater agreement, and Appendix~\ref{app:mfrm-calibration-sensitivity} examines sensitivity to the use of separate calibrations.

\subsection{Predicting Human-Easy and
LLM-Hard Cases}

To examine whether human--LLM mismatch is predictable from observable features, we formulate a binary classification task at the summary--dimension level. The positive class consists of units in the human-easy/LLM-hard quadrant, and all other units are treated as negative. We select features that capture observable properties of the source--summary pair. TF--IDF features capture lexical and topical content, length and compression features reflect information density and omission, and sentence counts provide coarse structural cues. Digit and negation counts approximate factual-detail and verification demands, while specific system IDs are excluded to avoid label leakage. We train class-balanced logistic regression models using these features. A detailed list of input features is provided in Appendix~\ref{app:input_features}.

We evaluate both a pooled model over all dimensions and dimension-specific models using 5-fold stratified cross-validation. In each fold, 80\% of the data are used for training and 20\% for testing while preserving the class ratio. Performance is reported using AUROC and AUPRC. Because the task is imbalanced, we interpret AUPRC relative to the positive-class rate.

%% file: 04results.tex
\section{Results}

\subsection{RQ1}

\paragraph{Model adequacy.}
Before comparing human and LLM judging structures, we verify that the MFRM estimations provide usable measurement models. Table~\ref{tab:mfrm-fit} summarizes the main fit diagnostics, including MAE and RMSE as descriptive summaries of rating-level prediction error, mean and standard deviation of standardized residuals, and item separation reliability. Across both estimations, standardized residuals are centered near zero and have standard deviations close to one. Item separation reliability is high in both estimations, indicating that the model distinguishes latent summary quality with sufficient precision. These diagnostics support using the fitted MFRM estimates to compare latent quality and residual hardness across human and LLM judge groups. All five response categories are used by both human and LLM raters, although their usage distributions differ.

\begin{table}[t]
\centering
\small
\setlength{\tabcolsep}{4pt}
\begin{tabular}{lrrrrr}
\toprule
Calib. & MAE & RMSE & Res. M & Res. SD & Item rel. \\
\midrule
Human & 0.712 & 0.892 & 0.005 & 1.015 & 0.859  \\
LLM   & 0.642 & 0.831 & 0.012 & 1.031 & 0.948  \\
\bottomrule
\end{tabular}
\caption{MFRM fit diagnostics. MAE and RMSE are descriptive summaries of rating-level prediction error. Res. M and Res. SD denote the mean and standard deviation of standardized residuals. Item rel. denotes item separation reliability.}
\label{tab:mfrm-fit}
\end{table}

\paragraph{Quality alignment does not imply hardness alignment.} At the summary level, LLM judges show moderate alignment with human latent summary quality. The correlation between human- and LLM MFRM quality estimates is Pearson \(r=0.372\) and Spearman \(\rho=0.371\) over 1,600 summaries. This indicates that LLM judges capture part of the human ranking of summary quality.

\begin{figure*}[t] 
\centering 
\includegraphics[width=\textwidth]{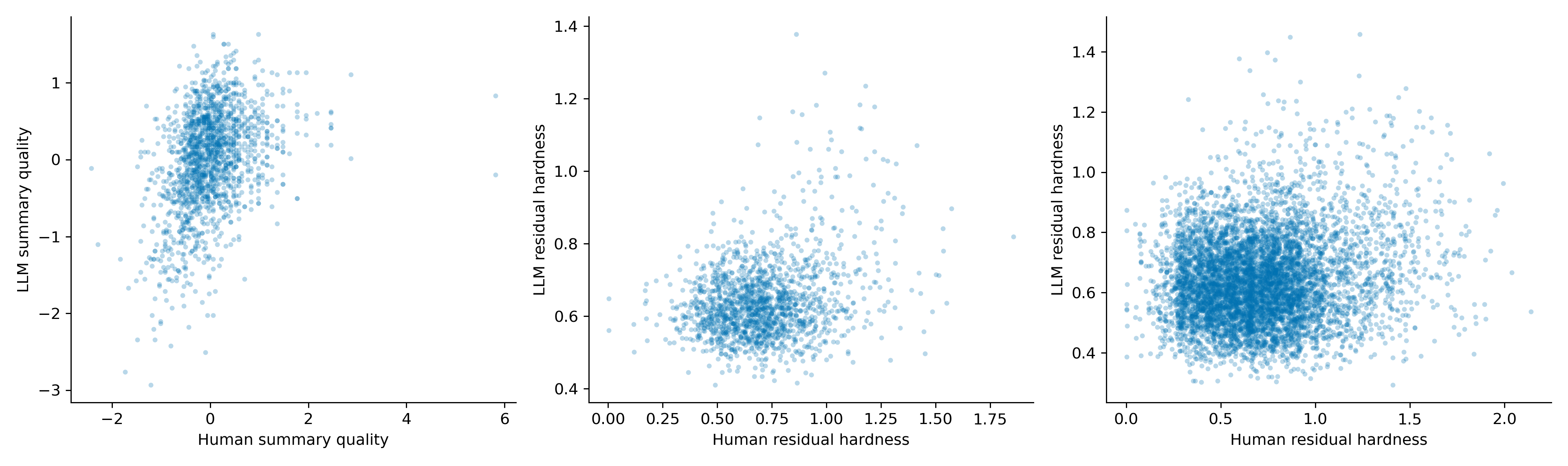} 
\caption{
Human--LLM alignment in latent summary quality (left), summary-level residual hardness (middle), and summary--dimension residual hardness (right).
}
\label{fig:quality-hardness-alignment} \end{figure*}

However, this alignment does not transfer to residual hardness. When residual hardness is aggregated at the same summary level, the human--LLM correlation drops to \(r=0.283\) and \(\rho=0.176\). The gap is even sharper at the summary-dimension level, where each point corresponds to a summary evaluated on one dimension. There, human and LLM residual hardness correlate only weakly, with \(r=0.145\) and \(\rho=0.089\) over 6,400 summary-dimension units. This pattern is also visible in Figure~\ref{fig:quality-hardness-alignment}. The quality estimates form a clear positive trend, whereas the hardness estimates are substantially more diffuse, especially at the summary--dimension level.

\paragraph{Robustness to judge capability and human-panel composition.}
The weak residual-hardness alignment is not driven solely by the smallest LLM judges. When we refit the LLM MFRM using only the seven judges with at least 7B parameters, quality alignment remains moderate (\(\rho=0.408\)), whereas summary--dimension residual-hardness alignment remains near zero (\(\rho=0.039\)). We additionally refit the human MFRM using only the three expert annotators and compare it with the independently fitted \(\geq7\)B LLM panel. In this setting, quality alignment is substantially stronger (\(\rho=0.688\)), while residual-hardness alignment remains weak (\(\rho=0.092\)). 
These analyses indicate that the contrast between quality alignment and residual-hardness alignment is not explained solely by the smallest LLM judges or by pooling expert and crowd human ratings.

These results answer RQ1 negatively. LLMs partially align with humans in latent summary quality, but this quality alignment does not carry over to the structure of residual hardness. In other words, LLMs can rank summaries in a moderately human-aligned way while exhibiting substantially different residual-hardness patterns.

\subsection{RQ2}

We next examine whether the weak overall residual-hardness alignment is accompanied by systematic differences across evaluation dimensions. We classified each summary--dimension unit as easy or hard within each judge group by thresholding residual hardness at the group's overall median. This median split is an operational definition rather than a theoretically privileged cutoff. Appendix~\ref{app:rq3-prediction-robustness} examines the sensitivity of the downstream RQ3 prediction analysis to a stricter quartile-based quadrant definition.

\begin{figure*}[t]
\centering
\includegraphics[width=0.65\textwidth]{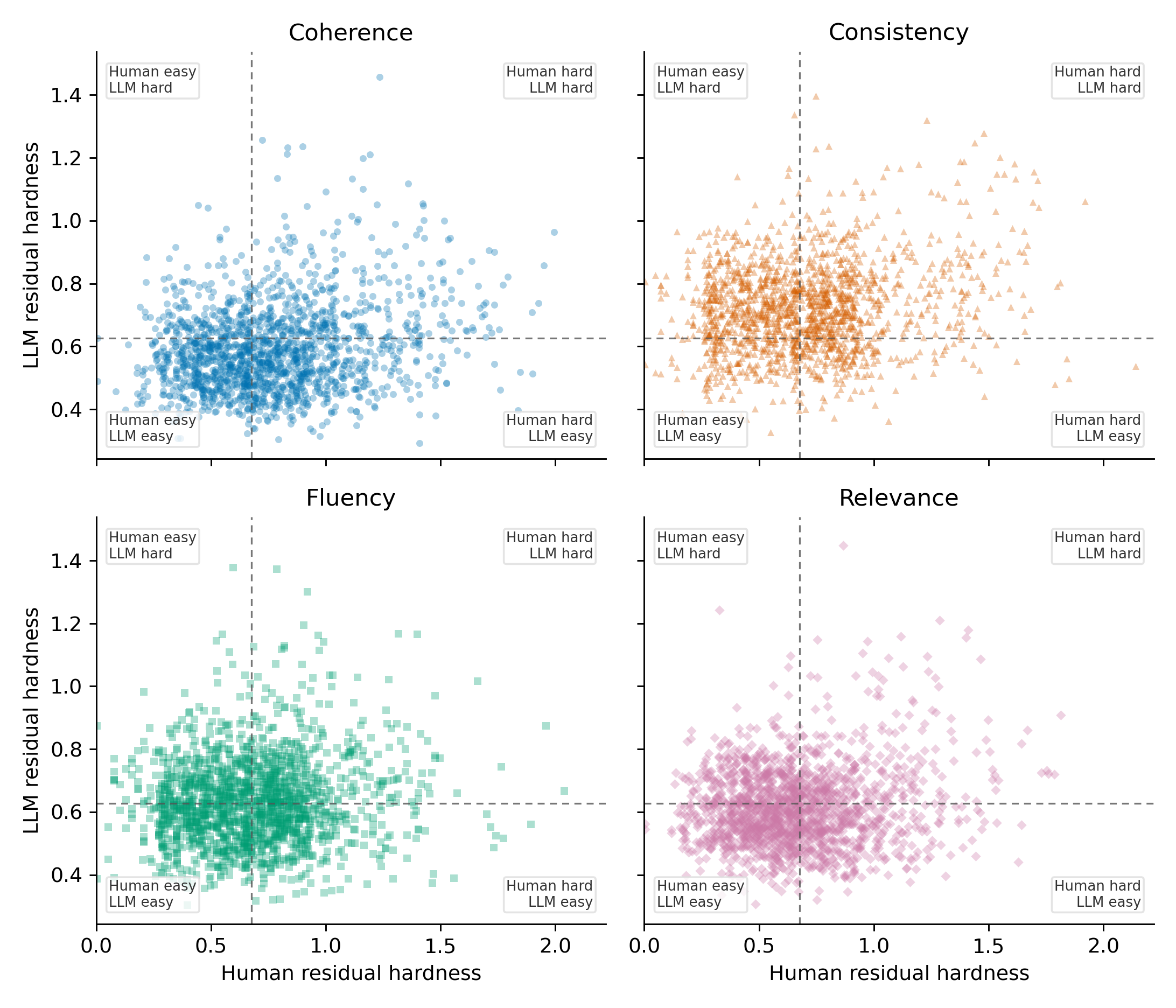}
\caption{
Human--LLM residual hardness by evaluation dimension. Dashed lines indicate the overall median residual hardness within each judge group, partitioning the space into human-easy/LLM-easy, human-easy/LLM-hard, human-hard/LLM-easy, and human-hard/LLM-hard regions.
}
\label{fig:quality-hardness-alignment-dim}
\end{figure*}

Table~\ref{tab:dimension-hardness-gap} reports the percentage of units classified as hard for each judge group. Consistency shows the largest LLM-hard shift, with 74.7\% of units classified as LLM-hard compared with 50.8\% human-hard. Coherence shows the opposite pattern, with 57.3\% human-hard but only 34.8\% LLM-hard. Fluency and relevance are closer to balanced, with much smaller hard-rate gaps. This structure is also visible in Figure~\ref{fig:quality-hardness-alignment-dim}, where consistency shows a pronounced concentration of points in the human-easy/LLM-hard region, whereas coherence shows the opposite tendency. This pattern also persists under the expert-only versus $\geq 7$B LLM comparison, where a \(+26.2\) percentage point consistency gap remains.

\begin{table}[t]
\centering
\small
\setlength{\tabcolsep}{4pt}
\begin{tabular}{lrrrr}
\toprule
Dimension & Human & LLM & Gap \\
\midrule
Consistency & 50.8 & 74.7 & +23.9 \\
Coherence   & 57.3 & 34.8 & -22.5  \\
Relevance   & 43.0 & 41.6 & -1.4  \\
Fluency     & 48.9 & 48.9 & -0.1  \\
\bottomrule
\end{tabular}
\caption{Dimension-level residual hardness gaps. Values are percentages of summary--dimension units classified as hard within each judge group. Gap is LLM minus human hard-rate.}
\label{tab:dimension-hardness-gap}
\end{table}

At the system-type level, the mismatch is weaker than at the dimension level. As reported in Table~\ref{tab:system-hardness-gap}, extractive systems show a modest LLM-hard shift, whereas abstractive systems show a small human-hard shift. 

We additionally examined rater-level fit and misfit sensitivity in Appendix~\ref{app:rater-fit}. The main residual-hardness findings remain stable after excluding raters flagged by descriptive infit or outfit criteria. To assess whether the conclusions depend on direct comparability of raw residual magnitudes across separate calibrations, we also report normalization- and calibration-based sensitivity checks in Appendix~\ref{app:mfrm-calibration-sensitivity}. The weak residual-hardness alignment and the dominant consistency/coherence contrast remain stable across these alternative specifications.

\begin{table}[t]
\centering
\small
\setlength{\tabcolsep}{4pt}
\begin{tabular}{lrrr}
\toprule
System type & Human & LLM & Gap \\
\midrule
Extractive  & 38.9 & 51.8 & +12.9 \\
Abstractive & 53.7 & 49.4 & -4.3  \\
\bottomrule
\end{tabular}
\caption{System-level residual hardness gaps. Values are percentages of summary--dimension units classified as hard. Gap is LLM minus human hard-rate.}
\label{tab:system-hardness-gap}
\end{table}

\subsection{RQ3}

For RQ3, we analyzed whether this mismatch can be predicted directly from observable properties of the source document and summary. Table~\ref{tab:llm-specific-hardness-prediction} shows that human-easy/LLM-hard cases are partially predictable from observable features. The pooled all-dimension model reaches AUROC 0.631 and AUPRC 0.341, compared with a positive rate of 23.3\%. Relevance is the most predictable task, with AUROC 0.717, followed by consistency, fluency, and coherence. These results suggest that the human-easy/LLM-hard quadrant is not arbitrary. It is associated with observable properties of the evaluated summary and source document. This predictive signal is robust to feature ablations and stricter grouped split strategies. Additional ablations, confidence intervals, grouped-split robustness checks and quadrant-definition sensitivity checks are reported in Appendix~\ref{app:rq3-prediction-robustness}.

\begin{table}[t]
\centering
\small
\begin{tabular}{lrrrr}
\toprule
Task & \(N\) & Pos. & AUROC & AUPRC \\
\midrule
All dimensions & 6,400 & 1,494 & 0.631 & 0.341 \\
Relevance & 1,600 & 356 & 0.717 & 0.442 \\
Consistency & 1,600 & 571 & 0.633 & 0.465 \\
Fluency & 1,600 & 374 & 0.631 & 0.353 \\
Coherence & 1,600 & 193 & 0.621 & 0.197 \\
\bottomrule
\end{tabular}
\caption{
Prediction of human-easy/LLM-hard cases from all summary-dimension units.
The positive class is the human-easy/LLM-hard quadrant; all other quadrants are negative.
}
\label{tab:llm-specific-hardness-prediction}
\end{table}

\begin{figure*}[t]
\centering
\includegraphics[width=\textwidth]{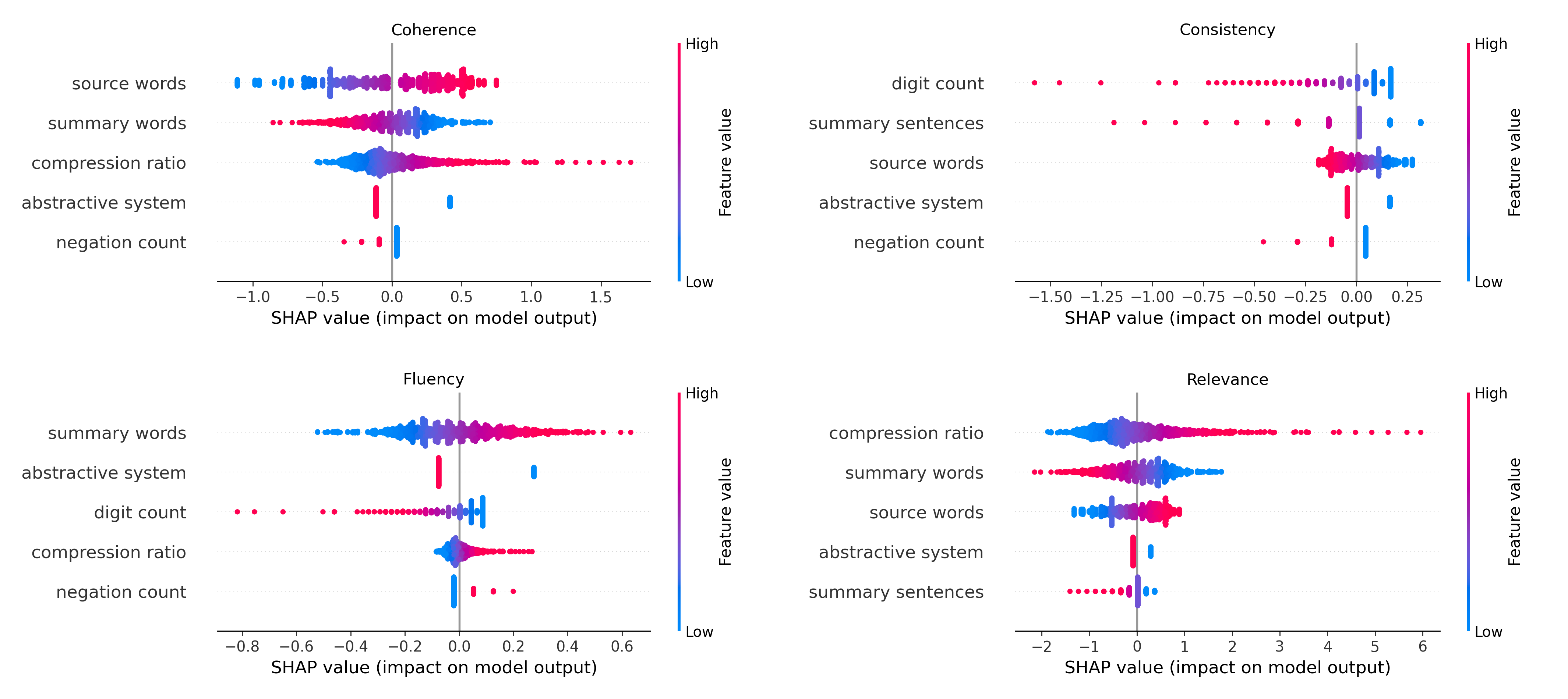}
\caption{
SHAP beeswarm plots for human-easy/LLM-hard classification by evaluation dimension. Positive SHAP values indicate features that push predictions toward the human-easy/LLM-hard quadrant, while negative values indicate the opposite. Color denotes feature value from low to high.
}
\label{fig:shap-dimension-prediction}
\end{figure*}

Figure~\ref{fig:shap-dimension-prediction} interprets the dimension-specific classifiers using signed SHAP values \cite{10.5555/3295222.3295230}. Because several surface features are correlated, particularly source length, summary length, and compression ratio, we interpret the SHAP values descriptively as model associations rather than as independent or causal feature effects. For relevance, higher compression ratio is associated with SHAP contributions toward human-easy/LLM-hard classification, whereas longer summaries tend to contribute in the opposite direction. Coherence shows a similar pattern, where longer source documents and higher compression ratios are associated with contributions toward the human-easy/LLM-hard quadrant, while longer summaries tend to contribute in the opposite direction.

Consistency exhibits a different profile. Its most influential predictors include digit count, sentence count, source length, negation count, and coarse system type, indicating that mismatch in this dimension is tied more strongly to factual-detail and verification-related cues than to compression alone. Fluency shows a weaker but still interpretable pattern, in which longer summaries contribute most strongly to human-easy/LLM-hard classification, with smaller contributions from system type, digit count, and compression ratio.

%% file: 05discusstion.tex
\section{Discussion}

Our results show that human alignment in LLM-as-judge evaluation is not a single property. Although LLM judges moderately align with human latent summary quality, this agreement does not imply that they share the same residual hardness structure. The weak correlation between human and LLM residual hardness indicates that the cases that remain difficult after controlling for summary quality, rater severity, dimension severity, and rating thresholds are only partially overlapping across judge groups. This distinction matters because aggregate quality alignment can make an evaluator appear reliable while masking item-level instability \cite{lalor-etal-2016-building,lalor-etal-2018-understanding}.

The mismatch structure appears at the dimension level. Consistency shows a large LLM-hard shift, whereas coherence shows the opposite human-hard shift. Consistency judgments require checking whether summary claims are supported by the source document \cite{fabbri-etal-2021-summeval,kryscinski-etal-2020-evaluating,wang-etal-2020-asking}, which may impose a factual-verification burden on LLM judges. In contrast, coherence judgments depend more on discourse organization, readability, and holistic interpretation \cite{zhao-etal-2023-discoscore, barzilay-lapata-2005-modeling, fabbri-etal-2021-summeval}, where human annotators may apply more nuanced or variable standards. Fluency and relevance show smaller gaps, suggesting that the divergence is not a general failure of the LLM judges but a dimension-specific difference in residual judging behavior.

The prediction results further indicate that part of the human--LLM residual-hardness mismatch is associated with observable source--summary properties, consistent with prior evidence that LLM-as-judge behavior exhibits persistent biases and dimension-dependent unreliability \cite{ye2025justice, shen-etal-2023-large}. In our setting, the class-balanced logistic regression model predicts human-easy/LLM-hard cases above the positive-class baseline, providing modest evidence that part of the observed mismatch is associated with observable source--summary properties. SHAP analysis reveals two feature profiles. For relevance and coherence, compression-related features dominate, providing modest evidence that some of the observed mismatch is predictable from observable source--summary properties. For consistency, digit count, sentence count, source length, negation count, and system type point to factual-detail and verification-related cues, consistent with prior work framing factual consistency evaluation as verification-intensive \cite{chen2023evaluatingfactualconsistencysummaries, gabriel-etal-2021-go}. Overall, part of the residual-hardness mismatch is measurable and partially predictable from observable text properties.

These findings have methodological implications for evaluating LLM judges. Existing validation practices \cite{liu2023geval,gu2025surveyllmasajudge} often treat human alignment as a score-level property, measured through correlations, rank agreement, or preference agreement. Our results show that such metrics are incomplete. Two judge groups may agree on the latent quality ordering of outputs while disagreeing on which individual cases are difficult to judge. Psychometric residual diagnostics therefore provide a complementary view of evaluator reliability. In particular, residual hardness and dimension-specific hardness gaps can reveal failure modes that are invisible to aggregate agreement metrics.

Practically, this suggests that LLM judges should not be deployed as uniform replacements for human annotators. Their reliability depends on the evaluation dimension and on properties of the evaluated text. A difficulty-aware evaluation pipeline could use such diagnostics to route cases selectively \cite{bondi2022role,10.1145/3613904.3641960}. Cases predicted to be human-easy but LLM-hard may be prioritized for human review, additional source-grounding checks, or stronger judge models. In this sense, these diagnostics may support more targeted human--LLM collaboration rather than simply accepting or rejecting LLM judges at the aggregate level.

%% file: 06conclution.tex
\section{Conclusion}

We presented a psychometric analysis of human--LLM alignment in summarization evaluation. Moving beyond aggregate agreement metrics, we asked whether human and LLM judges share the same structure of residual judging difficulty after controlling for latent summary quality, rater severity, dimension severity, and rating-scale thresholds. Using a Many-Facet Rasch Model, we showed that moderate human--LLM alignment in latent summary quality does not imply alignment in residual hardness. In particular, human and LLM judges differ substantially in their summary--dimension residual-hardness patterns, with especially pronounced mismatch for consistency and coherence.

We further showed that human-easy but LLM-hard cases are partially predictable from observable properties of the source--summary pair, with different predictive patterns across evaluation dimensions. These findings suggest that score-level human alignment provides only a partial view of LLM-as-judge reliability. Psychometric residual diagnostics can reveal failure modes that remain hidden under aggregate agreement metrics.

More broadly, our results argue for a difficulty-aware view of LLM-as-judge evaluation. Rather than treating LLM judges as uniform replacements for human annotators, future evaluation pipelines should account for which cases are likely to remain difficult for LLMs even when overall score alignment appears acceptable. This perspective can support more targeted human--LLM collaboration and more informative evaluation of automatic judges. The empirical findings are specific to the SummEval setting studied here, and establishing whether similar patterns generalize across tasks, domains, and stronger judge populations remains an important direction for future work.

%% file: 07limitation.tex
\section*{Limitations}

Our study has several limitations. First, we examine residual hardness mismatch in summarization evaluation using a single multi-rater benchmark, which limits the generalizability of the findings to other LLM-as-judge settings. However, this benchmark is well suited to our research question because it provides repeated ratings across multiple evaluation dimensions and includes anonymized rater identifiers required for the MFRM. More broadly, public summarization benchmarks with repeated human ratings, identifiable raters, and summary-level multidimensional quality scores are extremely limited, and SummEval is, to our knowledge, the most suitable publicly available benchmark for this analysis.

Second, our results are based on a fixed panel of open-weight LLM judges under a single prompting setup. Different model families, prompting strategies, or evaluation protocols may yield different residual hardness structures.

Third, residual hardness should not be interpreted as a standalone latent construct of judging difficulty or ambiguity. Rather, it is a post-fit diagnostic that operationalizes local judging difficulty through residual deviations from the fitted MFRM. Its magnitude may reflect local rater disagreement, model misspecification, or remaining structure not captured by the model. Accordingly, our analyses establish differences in residual-hardness patterns between the evaluated human and LLM panels, but do not assume that these residuals have identical substantive meaning across panels.

Fourth, the RQ3 prediction analysis identifies correlates of human--LLM mismatch from observable features, but does not establish the causal mechanisms underlying these patterns.

\section*{Acknowledgments} This research was conducted within the project “Assessment for Learning with AI (ALwAI)” funded by the Leibniz Association under the Leibniz Competition (project no. T163/2024).

%% file: 08appendix.tex
\section{Appendix}

\subsection{LLM Judge Prompt}
\label{app:llm-judge-prompt}

We use the same task instruction for all open-weight LLM judges. Table~\ref{tab:llm-judges} lists the 17 open-weight LLM judges used in our main analysis. For models whose tokenizer supports a system role, the prompt is formatted as a chat conversation using the model's native chat template. If the model does not support a system role, the system instruction is prepended to the user message. Decoding is deterministic.

\begin{quote}
\textbf{System:} You are a careful evaluator of news summaries.

\textbf{User:} Rate the summary for the requested evaluation dimension.

Use a 1--5 integer scale where 1 is very poor and 5 is excellent.

Dimension: \{\textit{dimension}\}

Criterion: \{\textit{rubric}\}

Source document:

\{\textit{source document}\}

Summary:

\{\textit{summary}\}

Return ONLY one number: 1, 2, 3, 4, or 5.

Answer:
\end{quote}

The dimension field is one of coherence, consistency, fluency, or relevance. The criterion field contains the corresponding SummEval rubric. We parse the first valid integer in \(\{1,\ldots,5\}\) from the model output. If no valid score is found, we retry once with a longer generation budget and apply the same parser. All prompts included the full source document without explicit truncation. As a length check, the longest rendered prompt contained 694 whitespace-delimited words, with maximum source and summary lengths of 540 and 133 words, respectively. These lengths are far below the context limits of the local judges.

\begin{table}[h]
\centering
\small
\begin{tabular}{lp{0.68\linewidth}}
\toprule
 Model \\
\midrule
 \texttt{Qwen/Qwen3.5-0.8B} \\
 \texttt{Qwen/Qwen3.5-2B} \\
\texttt{Qwen/Qwen3.5-4B} \\
 \texttt{Qwen/Qwen3.5-9B} \\
 \texttt{microsoft/Phi-4-mini-instruct} \\
\texttt{microsoft/Phi-3-mini-4k-instruct} \\
\texttt{microsoft/Phi-3-medium-4k-instruct} \\
\texttt{meta-llama/Llama-3.2-1B-Instruct} \\
\texttt{meta-llama/Llama-3.2-3B-Instruct} \\
\texttt{meta-llama/Llama-3.1-8B-Instruct} \\
\texttt{google/gemma-3-1b-it} \\
\texttt{google/gemma-3-4b-it} \\
\texttt{google/gemma-3-12b-it} \\
\texttt{mistralai/Ministral-3-3B-Instruct-2512} \\
\texttt{mistralai/Ministral-3-8B-Instruct-2512} \\
\texttt{mistralai/Ministral-3-14B-Instruct-2512} \\
\texttt{openchat/openchat-3.5-0106} \\
\bottomrule
\end{tabular}
\caption{17 open-weight LLM judges used for SummEval annotation in our analysis.}
\label{tab:llm-judges}
\end{table}

\subsection{MFRM}
\subsubsection{MFRM Estimation Details}
\label{app:mfrm_estimation}

We estimate the MFRM parameters by maximum likelihood. For each observed rater--summary--dimension rating, we compute \(\eta_{rid}=\theta_i-\rho_r-\delta_d\), construct the five category logits implied by the adjacent-category formulation, and normalize them with a softmax. The objective is the negative log-likelihood of the observed rating categories with weak \(L_2\) regularization. We optimize this objective in PyTorch using AdamW with learning rate \(0.05\) and weight decay \(10^{-4}\). For identification, summary, rater, and dimension parameters are mean-centered during optimization. The step thresholds are constrained to be ordered using an ordered parameterization and are initialized as evenly spaced values from \(-1.5\) to \(1.5\). All other facet parameters are initialized at zero. Optimization uses early stopping when the loss does not improve by \(10^{-7}\) for 300 epochs.

\subsubsection{Agreement-based Sanity Check for Residual Hardness}
\label{app:hardness_sanity_check}

As a sanity check, we examine whether residual hardness covaries with local rater agreement. For each summary--dimension unit \(i\), we compute the observed pairwise agreement
\begin{equation}
P_i =
\frac{\sum_c n_{ic}(n_{ic}-1)}
     {n_i(n_i-1)},
\end{equation}
where \(n_{ic}\) denotes the number of raters assigning category \(c\) to unit \(i\), and \(n_i\) is the total number of ratings for that unit. We then define the local chance-corrected agreement contribution as
\begin{equation}
\kappa_i =
\frac{P_i-P_e}{1-P_e},
\end{equation}
where \(P_e\) is estimated separately for the corresponding judge panel and evaluation dimension. This quantity is used as a local agreement diagnostic rather than as the standard global kappa statistic.

\begin{table}[t]
\centering
\small
\begin{tabular}{llcc}
\toprule
Group & Dimension & Exact \(\kappa\) & Weighted \(\kappa\) \\
\midrule
Human & Coherence & -0.547 & -0.854 \\
Human & Consistency & -0.481 & -0.865 \\
Human & Fluency & -0.434 & -0.855 \\
Human & Relevance & -0.645 & -0.828 \\
LLM & Coherence & -0.401 & -0.416 \\
LLM & Consistency & -0.374 & -0.615 \\
LLM & Fluency & -0.339 & -0.373 \\
LLM & Relevance & -0.385 & -0.494 \\
\bottomrule
\end{tabular}
\caption{Spearman correlations between residual hardness and summary--dimension-level agreement contribution. Negative correlations indicate that units with higher residual hardness contribute less to rater agreement.}
\label{tab:hardness_agreement_check}
\end{table}

Residual hardness is negatively associated with the local agreement diagnostics across all dimensions in both judge panels. This relationship is intended only as a sanity check showing that larger residual deviations tend to coincide with greater local rater disagreement. It does not independently validate residual hardness as a distinct construct or establish that residual hardness has identical substantive meaning across the human and LLM panels. Indeed, the associations are substantially stronger for human ratings than for LLM ratings. Unlike the agreement diagnostic, residual hardness is conditional on the fitted MFRM facets.

\subsubsection{Judge-Capability and Human-Panel Sensitivity}
\label{app:capable-judge-sensitivity}

\begin{table}[h]
\centering
\small
\setlength{\tabcolsep}{5pt}
\begin{tabular}{lcc}
\toprule
Panel comparison & Quality \(\rho\) & Residual \(\rho\) \\
\midrule
All human vs. all LLM & 0.371 & 0.089 \\
All human vs. \(\geq7\)B LLM & 0.408 & 0.039 \\
Experts vs. \(\geq7\)B LLM & 0.688 & 0.092 \\
\bottomrule
\end{tabular}
\caption{Sensitivity of human--LLM alignment to LLM judge capability and human-panel composition. Residual-hardness correlations are computed at the summary--dimension level.}
\label{tab:capable-judge-sensitivity}
\end{table}

Restricting the LLM panel to the seven judges with at least 7B parameters does not remove the contrast between quality alignment and residual-hardness alignment. The contrast is also preserved when the human MFRM is estimated using only the three expert annotators. In the expert-only versus \(\geq7\)B comparison, the consistency hard-rate gap remains \(+26.2\) percentage points. These results indicate that the main RQ1 and RQ2 patterns are not driven solely by the smallest LLM judges or by expert--crowd disagreement in the human panel.

\subsubsection{Rater-by-Dimension Interaction Sensitivity}
\label{app:rater-dimension-interaction}

To examine whether the weak human--LLM residual-hardness alignment primarily reflects an omitted rater-by-dimension interaction, we refit the MFRM with an additional rater--dimension term. Under this richer specification, human--LLM residual-hardness alignment remains weak (\(r=0.277\), \(\rho=0.122\)). Thus, the main RQ1 pattern is not solely attributable to omission of rater-specific dimension effects.

\subsubsection{Rater-Misfit Sensitivity}
\label{app:rater-fit}

To examine whether the main residual-hardness findings are driven by poorly fitting raters, we conducted a post-hoc sensitivity analysis after excluding raters with infit or outfit mean-square outside the conventional descriptive range of 0.5--1.5. The original MFRM calibrations were kept fixed, and residual-hardness summaries were recomputed using only the remaining raters.

Table~\ref{tab:misfit-exclusion-rq1} shows that the central RQ1 conclusion remains unchanged. Human--LLM alignment in latent quality remains moderate, whereas residual-hardness alignment remains weak at both the summary level and the summary--dimension level. If anything, excluding flagged raters further reduces the residual-hardness correlations, suggesting that the weak human--LLM hardness correspondence is not an artifact of poorly fitting raters.

\begin{table}[h]
\centering
\small
\setlength{\tabcolsep}{6pt}
\begin{tabular}{lccc}
\toprule
Outcome & Stat. & All raters & Exclude flagged \\
\midrule
Quality & \(r\) & 0.372 & 0.372 \\
 & \(\rho\) & 0.371 & 0.371 \\
Summary hardness & \(r\) & 0.283 & 0.222 \\
 & \(\rho\) & 0.176 & 0.157 \\
Dim. hardness & \(r\) & 0.145 & 0.108 \\
 & \(\rho\) & 0.089 & 0.073 \\
\bottomrule
\end{tabular}
\caption{RQ1 sensitivity after excluding raters with infit or outfit mean-square outside the descriptive 0.5--1.5 range.}
\label{tab:misfit-exclusion-rq1}
\end{table}

Table~\ref{tab:misfit-exclusion-rq2} shows that the dimension-level pattern is also stable. Consistency remains LLM-hard relative to humans, whereas coherence remains human-hard relative to LLMs. The smaller gaps for fluency and relevance vary more after exclusion, which is expected because their original gaps are close to zero.

\begin{table}[h]
\centering
\small
\setlength{\tabcolsep}{4pt}
\begin{tabular}{lrr}
\toprule
Dimension & All raters gap & Exclude flagged gap \\
\midrule
Coherence & -22.5 & -19.8 \\
Consistency & 23.9 & 18.0 \\
Fluency & -0.1 & 7.3 \\
Relevance & -1.4 & -5.5 \\
\bottomrule
\end{tabular}
\caption{RQ2 dimension hard-rate gap sensitivity after excluding flagged raters. Values are percentage-point gaps, computed as LLM hard-rate minus human hard-rate.}
\label{tab:misfit-exclusion-rq2}
\end{table}

\subsubsection{MFRM Calibration Comparability Sensitivity}
\label{app:mfrm-calibration-sensitivity}

Because the main analyses compare residual hardness derived from separately calibrated human-only and LLM-only MFRMs, we examine whether the conclusions depend on direct comparability of raw residual magnitudes across calibrations. We recompute the residual-hardness comparisons under three alternatives: within-group \(z\)-scored residual hardness, percentile-rank residual hardness, and residuals from a combined MFRM fit to pooled human and LLM ratings. For the combined-model analysis, absolute residuals are aggregated separately for human and LLM raters after fitting the shared model.

\begin{table}[h]
\centering
\small
\setlength{\tabcolsep}{6pt}
\begin{tabular}{llcc}
\toprule
Method & Stat. & Summary & Dim. \\
\midrule
Main & \(r\) & 0.283 & 0.145 \\
 & \(\rho\) & 0.176 & 0.089 \\
\(z\)-score & \(r\) & 0.283 & 0.145 \\
 & \(\rho\) & 0.176 & 0.089 \\
Rank & \(r\) & 0.176 & 0.089 \\
 & \(\rho\) & 0.176 & 0.089 \\
Combined & \(r\) & 0.352 & 0.170 \\
 & \(\rho\) & 0.231 & 0.104 \\
\bottomrule
\end{tabular}
\caption{Residual-hardness alignment under alternative calibration and normalization choices.}
\label{tab:mfrm-calibration-sensitivity-correlations}
\end{table}

The main conclusion does not rely on direct comparability of raw residual magnitudes across separately calibrated models. As expected, within-group \(z\)-scoring leaves the correlations unchanged because it preserves within-group ordering. Rank-based normalization yields slightly smaller correlations, but the overall pattern remains the same. Under the combined MFRM, residual-hardness alignment is somewhat higher, yet it remains clearly weaker than the corresponding quality alignment and remains weak in absolute terms at the summary--dimension level. The dominant dimension pattern is also stable across specifications. Consistency remains LLM-hard relative to humans, whereas coherence remains human-hard relative to LLMs.

\subsection{Additional Information and Experiments for RQ3}

\subsubsection{Input Features for RQ3}
\label{app:input_features}
Table~\ref{tab:rq3-prediction-features} summarizes the input features used for the RQ3 human-easy/LLM-hard prediction experiments.
We include these features as interpretable, observable proxies for factors that may affect judging difficulty. TF--IDF features capture lexical and topical content, length and compression features approximate information density and omission, and sentence, digit, and negation counts provide coarse structural and factual-detail cues. Coarse system type captures broad extractive versus abstractive differences, while specific system IDs are excluded to reduce label leakage.

\begin{table}[h]
\centering
\small
\begin{tabular}{p{0.22\linewidth}p{0.68\linewidth}}
\toprule
Feature group & Features \\
\midrule
Summary text & TF--IDF features from the generated summary, using unigram and bigram features with English stop-word removal. \\
Source text & TF--IDF features from the source document, using unigram and bigram features with English stop-word removal. \\
Surface features & Summary word count, source word count, compression ratio, summary sentence count, summary digit count, and summary negation count. \\
Coarse metadata & System type, coded as extractive or abstractive. \\
\bottomrule
\end{tabular}
\caption{Input features used for RQ3 prediction of human-easy/LLM-hard cases. TF--IDF features use \(\texttt{min\_df}=2\), unigram/bigram ranges, and up to 6,000 features each for the summary and source fields.}
\label{tab:rq3-prediction-features}
\end{table}

\subsubsection{RQ3 Prediction Robustness and Ablations}
\label{app:rq3-prediction-robustness}

We run additional robustness checks for the RQ3 human-easy/LLM-hard prediction task. All analyses use the pooled summary--dimension dataset (\(N=6{,}400\), positive class \(=1{,}494\), positive rate \(=23.3\%\)). As in the main analysis, we retain coarse system type but exclude specific system IDs. Confidence intervals are percentile bootstrap intervals computed from out-of-fold predictions.

\begin{table}[h]
\centering
\small
\setlength{\tabcolsep}{4pt}
\begin{tabular}{lcc}
\toprule
Feature set & AUROC & AUPRC \\
\midrule
Surface only & 0.573 \,[0.555, 0.590] & 0.293 \,[0.273, 0.314] \\
TF--IDF only & 0.620 \,[0.605, 0.637] & 0.326 \,[0.305, 0.351] \\
Metadata only & 0.542 \,[0.524, 0.560] & 0.260 \,[0.245, 0.277] \\
Full model & 0.631 \,[0.614, 0.648] & 0.341 \,[0.319, 0.365] \\
\bottomrule
\end{tabular}
\caption{RQ3 feature ablation for predicting human-easy/LLM-hard summary--dimension units. Brackets denote 95\% bootstrap confidence intervals from out-of-fold predictions. Metadata includes coarse system type only; specific system IDs are excluded.}
\label{tab:rq3-prediction-ablation}
\end{table}

\begin{table}[h]
\centering
\small
\setlength{\tabcolsep}{2pt}
\begin{tabular}{lcc}
\toprule
Split strategy & AUROC & AUPRC \\
\midrule
Stratified 5-fold & 0.631 \,[0.614, 0.648] & 0.341 \,[0.319, 0.365] \\
Summary & 0.618 \,[0.602, 0.634] & 0.326 \,[0.306, 0.348] \\
Document & 0.569 \,[0.553, 0.586] & 0.289 \,[0.269, 0.310] \\
System & 0.618 \,[0.602, 0.634] & 0.329 \,[0.309, 0.353] \\
\bottomrule
\end{tabular}
\caption{RQ3 split robustness for the full model. Brackets denote 95\% bootstrap confidence intervals from out-of-fold predictions. Grouped splits hold out all dimensions of the same summary, all summaries from the same source document, or all summaries from the same generation system, respectively.}
\label{tab:rq3-prediction-split-robustness}
\end{table}

These checks show that the RQ3 signal is not driven by a single feature family or solely by random-fold leakage across repeated summaries. TF--IDF features provide most of the predictive signal, surface features retain above-chance predictive performance, and the full model consistently improves over the ablations. Performance decreases under stricter grouped splits, especially when holding out entire source documents, but predictive signal remains in all settings. For the surface-feature model, we additionally perform a compression-drop ablation under summary-grouped 5-fold cross-validation. Removing compression ratio changes AUROC only from $0.571$ to $0.566$ and AUPRC from $0.293$ to $0.286$. Overall, these results support partial predictability of human-easy/LLM-hard cases that persists under alternative feature sets and split strategies rather than high-accuracy classification.

\begin{table}[h]
\centering
\small
\setlength{\tabcolsep}{5pt}
\begin{tabular}{lrrrr}
\toprule
Task & \(N\) & Pos. & AUROC & AUPRC \\
\midrule
All dimensions & 6,400 & 349 & 0.676 & 0.114 \\
Relevance & 1,600 & 102 & 0.703 & 0.206 \\
Consistency & 1,600 & 87 & 0.672 & 0.133 \\
Fluency & 1,600 & 79 & 0.743 & 0.138 \\
Coherence & 1,600 & 70 & 0.578 & 0.067 \\
\bottomrule
\end{tabular}
\caption{RQ3 quartile-threshold sensitivity for predicting human-easy/LLM-hard cases from all summary--dimension units.}
\label{tab:rq3-quartile-threshold-sensitivity}
\end{table}

We also test a stricter quadrant definition based on task-specific quartiles. For each prediction task, human-easy units are those whose human residual hardness falls in the lower quartile of that task's evaluation group, and LLM-hard units are those whose LLM residual hardness falls in the upper quartile of the same group. Table~\ref{tab:rq3-quartile-threshold-sensitivity} shows that the pooled predictive signal remains above chance under this more selective definition (AUROC \(=0.676\)), although dimension-specific robustness is uneven. In particular, coherence decreases to AUROC \(=0.578\). We therefore interpret this analysis as supporting pooled predictability rather than uniformly strong robustness across dimensions.

%% file: custom.bib
@inproceedings{lalor-etal-2016-building,
    title = "Building an Evaluation Scale using Item Response Theory",
    author = "Lalor, John P.  and
      Wu, Hao  and
      Yu, Hong",
    editor = "Su, Jian  and
      Duh, Kevin  and
      Carreras, Xavier",
    booktitle = "Proceedings of the 2016 Conference on Empirical Methods in Natural Language Processing",
    month = nov,
    year = "2016",
    address = "Austin, Texas",
    publisher = "Association for Computational Linguistics",
    url = "https://aclanthology.org/D16-1062/",
    doi = "10.18653/v1/D16-1062",
    pages = "648--657"
}

@inproceedings{lalor-etal-2018-understanding,
    title = "Understanding Deep Learning Performance through an Examination of Test Set Difficulty: A Psychometric Case Study",
    author = "Lalor, John P.  and
      Wu, Hao  and
      Munkhdalai, Tsendsuren  and
      Yu, Hong",
    editor = "Riloff, Ellen  and
      Chiang, David  and
      Hockenmaier, Julia  and
      Tsujii, Jun{'}ichi",
    booktitle = "Proceedings of the 2018 Conference on Empirical Methods in Natural Language Processing",
    month = oct # "-" # nov,
    year = "2018",
    address = "Brussels, Belgium",
    publisher = "Association for Computational Linguistics",
    url = "https://aclanthology.org/D18-1500/",
    doi = "10.18653/v1/D18-1500",
    pages = "4711--4716"
}

@inproceedings{zhou2025lost,
author = {Zhou, Hongli and Huang, Hui and Zhao, Ziqing and Han, Lvyuan and Wang, Huicheng and Chen, Kehai and Yang, Muyun and Bao, Wei and Dong, Jian and Xu, Bing and Zhu, Conghui and Cao, Hailong and Zhao, Tiejun},
title = {Lost in benchmarks? rethinking large language model benchmarking with item response theory},
year = {2026},
isbn = {978-1-57735-906-7},
publisher = {AAAI Press},
url = {https://doi.org/10.1609/aaai.v40i41.40814},
doi = {10.1609/aaai.v40i41.40814},
booktitle = {Proceedings of the Fortieth AAAI Conference on Artificial Intelligence and Thirty-Eighth Conference on Innovative Applications of Artificial Intelligence and Sixteenth Symposium on Educational Advances in Artificial Intelligence},
articleno = {3913},
numpages = {9},
series = {AAAI'26/IAAI'26/EAAI'26}
}

@misc{gu2025surveyllmasajudge,
      title={A Survey on LLM-as-a-Judge}, 
      author={Jiawei Gu and Xuhui Jiang and Zhichao Shi and Hexiang Tan and Xuehao Zhai and Chengjin Xu and Wei Li and Yinghan Shen and Shengjie Ma and Honghao Liu and Saizhuo Wang and Kun Zhang and Yuanzhuo Wang and Wen Gao and Lionel Ni and Jian Guo},
      year={2025},
      eprint={2411.15594},
      archivePrefix={arXiv},
      primaryClass={cs.CL},
      url={https://arxiv.org/abs/2411.15594}, 
}

@inproceedings{Cong2026ASAG,
author = {Cong, Longwei and Hammerla, Leon and Hahn, Sonja and Gombert, Sebastian and Drachsler, Hendrik and Kroehne, Ulf},
title = {Automatic Short Answer Grading with {LLMs}: From Memorization to Reasoning},
year = {2026},
publisher = {Association for Computing Machinery},
address = {New York, NY, USA},
doi = {10.1145/3785022.3785031},
booktitle = {Proceedings of the 16th International Learning Analytics and Knowledge Conference}
}

@book{van2016handbook,
  title={Handbook of Item Response Theory: Volume 1: Models},
  editor={van der Linden, W.J.},
  isbn={9781315374512},
  series={Behavioral Sciences},
  doi={10.1201/9781315374512},
  year={2016},
  publisher={Chapman and Hall/CRC}
}

@article{choi2026diagnosing,
  title={Diagnosing the Reliability of LLM-as-a-Judge via Item Response Theory},
  author={Choi, Junhyuk and Park, Sohhyung and Cho, Chanhee and Park, Hyeonchu and Kim, Bugeun},
  journal={arXiv preprint arXiv:2602.00521},
  year={2026},
  doi={10.48550/arXiv.2602.00521}
}

@inproceedings{rodriguez-etal-2021-evaluation,
    title = "Evaluation Examples are not Equally Informative: How should that change {NLP} Leaderboards?",
    author = "Rodriguez, Pedro  and
      Barrow, Joe  and
      Hoyle, Alexander  and
      Lalor, John P.  and
      Jia, Robin  and
      Boyd-Graber, Jordan",
    editor = "Zong, Chengqing  and
      Xia, Fei  and
      Li, Wenjie  and
      Navigli, Roberto",
    booktitle = "Proceedings of the 59th Annual Meeting of the Association for Computational Linguistics and the 11th International Joint Conference on Natural Language Processing (Volume 1: Long Papers)",
    month = aug,
    year = "2021",
    address = "Online",
    publisher = "Association for Computational Linguistics",
    url = "https://aclanthology.org/2021.acl-long.346/",
    doi = "10.18653/v1/2021.acl-long.346",
    pages = "4486--4503"
}

@inproceedings{gombert2026rubrics,
author = {Gombert, Sebastian and Sun, Zhifan and Zehner, Fabian and Lossjew, Jannik and Wyrwich, Tobias and Czinczel, Berrit Katharina and Bednorz, David and Kubsch, Marcus and Di Mitri, Daniele and Neumann, Knut and Drachsler, Hendrik},
title = {Are rubrics all you need? Towards rubric-based automatic short answer scoring via guided rubric-answer alignment},
year = {2026},
isbn = {9798400720666},
publisher = {Association for Computing Machinery},
address = {New York, NY, USA},
url = {https://doi.org/10.1145/3785022.3785064},
doi = {10.1145/3785022.3785064},
booktitle = {Proceedings of the LAK26: 16th International Learning Analytics and Knowledge Conference},
pages = {272–282},
numpages = {11},
location = {
},
series = {LAK '26}
}

@inproceedings{li-etal-2025-generation,
    title = "From Generation to Judgment: Opportunities and Challenges of {LLM}-as-a-judge",
    author = "Li, Dawei  and
      Jiang, Bohan  and
      Huang, Liangjie  and
      Beigi, Alimohammad  and
      Zhao, Chengshuai  and
      Tan, Zhen  and
      Bhattacharjee, Amrita  and
      Jiang, Yuxuan  and
      Chen, Canyu  and
      Wu, Tianhao  and
      Shu, Kai  and
      Cheng, Lu  and
      Liu, Huan",
    editor = "Christodoulopoulos, Christos  and
      Chakraborty, Tanmoy  and
      Rose, Carolyn  and
      Peng, Violet",
    booktitle = "Proceedings of the 2025 Conference on Empirical Methods in Natural Language Processing",
    month = nov,
    year = "2025",
    address = "Suzhou, China",
    publisher = "Association for Computational Linguistics",
    url = "https://aclanthology.org/2025.emnlp-main.138/",
    doi = "10.18653/v1/2025.emnlp-main.138",
    pages = "2757--2791",
    ISBN = "979-8-89176-332-6"
}

@inproceedings{tan2024large,
    title = "Large Language Models for Data Annotation and Synthesis: A Survey",
    author = "Tan, Zhen  and
      Li, Dawei  and
      Wang, Song  and
      Beigi, Alimohammad  and
      Jiang, Bohan  and
      Bhattacharjee, Amrita  and
      Karami, Mansooreh  and
      Li, Jundong  and
      Cheng, Lu  and
      Liu, Huan",
    editor = "Al-Onaizan, Yaser  and
      Bansal, Mohit  and
      Chen, Yun-Nung",
    booktitle = "Proceedings of the 2024 Conference on Empirical Methods in Natural Language Processing",
    month = nov,
    year = "2024",
    address = "Miami, Florida, USA",
    publisher = "Association for Computational Linguistics",
    url = "https://aclanthology.org/2024.emnlp-main.54/",
    doi = "10.18653/v1/2024.emnlp-main.54",
    pages = "930--957"
}

@inproceedings{liu2023geval,
    title = "{G}-Eval: {NLG} Evaluation using Gpt-4 with Better Human Alignment",
    author = "Liu, Yang  and
      Iter, Dan  and
      Xu, Yichong  and
      Wang, Shuohang  and
      Xu, Ruochen  and
      Zhu, Chenguang",
    editor = "Bouamor, Houda  and
      Pino, Juan  and
      Bali, Kalika",
    booktitle = "Proceedings of the 2023 Conference on Empirical Methods in Natural Language Processing",
    month = dec,
    year = "2023",
    address = "Singapore",
    publisher = "Association for Computational Linguistics",
    url = "https://aclanthology.org/2023.emnlp-main.153/",
    doi = "10.18653/v1/2023.emnlp-main.153",
    pages = "2511--2522"
}

@inproceedings{fu2024gptscore,
    title = "{GPTS}core: Evaluate as You Desire",
    author = "Fu, Jinlan  and
      Ng, See-Kiong  and
      Jiang, Zhengbao  and
      Liu, Pengfei",
    editor = "Duh, Kevin  and
      Gomez, Helena  and
      Bethard, Steven",
    booktitle = "Proceedings of the 2024 Conference of the North American Chapter of the Association for Computational Linguistics: Human Language Technologies (Volume 1: Long Papers)",
    month = jun,
    year = "2024",
    address = "Mexico City, Mexico",
    publisher = "Association for Computational Linguistics",
    url = "https://aclanthology.org/2024.naacl-long.365/",
    doi = "10.18653/v1/2024.naacl-long.365",
    pages = "6556--6576"
}

@inproceedings{zheng2023judging,
author = {Zheng, Lianmin and Chiang, Wei-Lin and Sheng, Ying and Zhuang, Siyuan and Wu, Zhanghao and Zhuang, Yonghao and Lin, Zi and Li, Zhuohan and Li, Dacheng and Xing, Eric P. and Zhang, Hao and Gonzalez, Joseph E. and Stoica, Ion},
title = {Judging LLM-as-a-judge with MT-bench and Chatbot Arena},
year = {2023},
publisher = {Curran Associates Inc.},
address = {Red Hook, NY, USA},
booktitle = {Proceedings of the 37th International Conference on Neural Information Processing Systems},
articleno = {2020},
numpages = {29},
location = {New Orleans, LA, USA},
series = {NIPS '23}
}

@inproceedings{wang2024pandalm,
 author = {Wang, Yidong and Yu, Zhuohao and Yao, Wenjin and Zeng, Zhengran and Yang, Linyi and Wang, Cunxiang and Chen, Hao and Jiang, Chaoya and Xie, Rui and Wang, Jindong and Xie, Xing and Ye, Wei and Zhang, Shikun and Zhang, Yue},
 booktitle = {International Conference on Learning Representations},
 editor = {B. Kim and Y. Yue and S. Chaudhuri and K. Fragkiadaki and M. Khan and Y. Sun},
 pages = {43573--43593},
 title = {PandaLM: An Automatic Evaluation Benchmark for LLM Instruction Tuning Optimization},
 url = {https://proceedings.iclr.cc/paper_files/paper/2024/file/be3b0d51a2b86cb4ffe50f13480217e0-Paper-Conference.pdf},
 volume = {2024},
 year = {2024}
}

@misc{li2023alpacaeval,
  author = {Xuechen Li and Tianyi Zhang and Yann Dubois and Rohan Taori and Ishaan Gulrajani and Carlos Guestrin and Percy Liang and Tatsunori B. Hashimoto },
  title = {AlpacaEval: An Automatic Evaluator of Instruction-following Models},
  year = {2023},
  month = {5},
  publisher = {GitHub},
  journal = {GitHub repository},
  howpublished = {\url{https://github.com/tatsu-lab/alpaca_eval}}
}

@inproceedings{bavaresco2025llms,
    title = "{LLM}s instead of Human Judges? A Large Scale Empirical Study across 20 {NLP} Evaluation Tasks",
    author = "Bavaresco, Anna  and
      Bernardi, Raffaella  and
      Bertolazzi, Leonardo  and
      Elliott, Desmond  and
      Fern{\'a}ndez, Raquel  and
      Gatt, Albert  and
      Ghaleb, Esam  and
      Giulianelli, Mario  and
      Hanna, Michael  and
      Koller, Alexander  and
      Martins, Andre  and
      Mondorf, Philipp  and
      Neplenbroek, Vera  and
      Pezzelle, Sandro  and
      Plank, Barbara  and
      Schlangen, David  and
      Suglia, Alessandro  and
      Surikuchi, Aditya K  and
      Takmaz, Ece  and
      Testoni, Alberto",
    editor = "Che, Wanxiang  and
      Nabende, Joyce  and
      Shutova, Ekaterina  and
      Pilehvar, Mohammad Taher",
    booktitle = "Proceedings of the 63rd Annual Meeting of the Association for Computational Linguistics (Volume 2: Short Papers)",
    month = jul,
    year = "2025",
    address = "Vienna, Austria",
    publisher = "Association for Computational Linguistics",
    url = "https://aclanthology.org/2025.acl-short.20/",
    doi = "10.18653/v1/2025.acl-short.20",
    pages = "238--255",
    ISBN = "979-8-89176-252-7"
}

@book { eckes2023introduction,
    title = "Introduction to Many-Facet Rasch Measurement",
    subtitle = "Analyzing and Evaluating Rater-Mediated Assessments. 2nd Revised and Updated Edition",
    author = "Thomas {Eckes}",
    year = "2015",
    publisher = "Peter Lang Verlag",
    address = "Berlin, Deutschland",
    doi = "10.3726/978-3-653-04844-5",
    url = "https://www.peterlang.com/document/1045610"
}

@misc{jiao2025comparinghumanairater,
      title={Comparing Human and AI Rater Effects Using the Many-Facet Rasch Model}, 
      author={Hong Jiao and Dan Song and Won-Chan Lee},
      year={2025},
      eprint={2505.18486},
      archivePrefix={arXiv},
      primaryClass={cs.CL},
      url={https://arxiv.org/abs/2505.18486}, 
}

@article{WANG2025100481,
title = {Evaluating large language models as raters in large-scale writing assessments: A psychometric framework for reliability and validity},
journal = {Computers and Education: Artificial Intelligence},
volume = {9},
pages = {100481},
year = {2025},
issn = {2666-920X},
doi = {https://doi.org/10.1016/j.caeai.2025.100481},
url = {https://www.sciencedirect.com/science/article/pii/S2666920X25001213},
author = {Yuehan Wang and Jinyan Huang and Lun Du and Yuxin Guo and Ying Liu and Rong Wang},
}

@article{fabbri-etal-2021-summeval,
    title = "{S}umm{E}val: Re-evaluating Summarization Evaluation",
    author = "Fabbri, Alexander R.  and
      Kry{\'s}ci{\'n}ski, Wojciech  and
      McCann, Bryan  and
      Xiong, Caiming  and
      Socher, Richard  and
      Radev, Dragomir",
    editor = "Roark, Brian  and
      Nenkova, Ani",
    journal = "Transactions of the Association for Computational Linguistics",
    volume = "9",
    year = "2021",
    address = "Cambridge, MA",
    publisher = "MIT Press",
    url = "https://aclanthology.org/2021.tacl-1.24/",
    doi = "10.1162/tacl_a_00373",
    pages = "391--409"
}

@article{bondi2022role, title={Role of Human-AI Interaction in Selective Prediction}, volume={36}, url={https://ojs.aaai.org/index.php/AAAI/article/view/20465}, DOI={10.1609/aaai.v36i5.20465}, abstractNote={Recent work has shown the potential benefit of selective prediction systems that can learn to defer to a human when the predictions of the AI are unreliable, particularly to improve the reliability of AI systems in high-stakes applications like healthcare or conservation. However, most prior work assumes that human behavior remains unchanged when they solve a prediction task as part of a human-AI team as opposed to by themselves. We show that this is not the case by performing experiments to quantify human-AI interaction in the context of selective prediction. In particular, we study the impact of communicating different types of information to humans about the AI system’s decision to defer. Using real-world conservation data and a selective prediction system that improves expected accuracy over that of the human or AI system working individually, we show that this messaging has a significant impact on the accuracy of human judgements. Our results study two components of the messaging strategy: 1) Whether humans are informed about the prediction of the AI system and 2) Whether they are informed about the decision of the selective prediction system to defer. By manipulating these messaging components, we show that it is possible to significantly boost human performance by informing the human of the decision to defer, but not revealing the prediction of the AI. We therefore show that it is vital to consider how the decision to defer is communicated to a human when designing selective prediction systems, and that the composite accuracy of a human-AI team must be carefully evaluated using a human-in-the-loop framework.}, number={5}, journal={Proceedings of the AAAI Conference on Artificial Intelligence}, author={Bondi, Elizabeth and Koster, Raphael and Sheahan, Hannah and Chadwick, Martin and Bachrach, Yoram and Cemgil, Taylan and Paquet, Ulrich and Dvijotham, Krishnamurthy}, year={2022}, month={Jun.}, pages={5286–5294} }

@inproceedings{10.1145/3613904.3641960,
author = {Wang, Xinru and Kim, Hannah and Rahman, Sajjadur and Mitra, Kushan and Miao, Zhengjie},
title = {Human-LLM Collaborative Annotation Through Effective Verification of LLM Labels},
year = {2024},
isbn = {9798400703300},
publisher = {Association for Computing Machinery},
address = {New York, NY, USA},
url = {https://doi.org/10.1145/3613904.3641960},
doi = {10.1145/3613904.3641960},
booktitle = {Proceedings of the 2024 CHI Conference on Human Factors in Computing Systems},
articleno = {303},
numpages = {21},
location = {Honolulu, HI, USA},
series = {CHI '24}
}

@inproceedings{wang-etal-2024-sotopia,
    title = "{SOTOPIA}-{\ensuremath{\pi}}: Interactive Learning of Socially Intelligent Language Agents",
    author = "Wang, Ruiyi  and
      Yu, Haofei  and
      Zhang, Wenxin  and
      Qi, Zhengyang  and
      Sap, Maarten  and
      Bisk, Yonatan  and
      Neubig, Graham  and
      Zhu, Hao",
    editor = "Ku, Lun-Wei  and
      Martins, Andre  and
      Srikumar, Vivek",
    booktitle = "Proceedings of the 62nd Annual Meeting of the Association for Computational Linguistics (Volume 1: Long Papers)",
    month = aug,
    year = "2024",
    address = "Bangkok, Thailand",
    publisher = "Association for Computational Linguistics",
    url = "https://aclanthology.org/2024.acl-long.698/",
    doi = "10.18653/v1/2024.acl-long.698",
    pages = "12912--12940"
}

@inproceedings{chen2024mllm,
author = {Chen, Dongping and Chen, Ruoxi and Zhang, Shilin and Wang, Yaochen and Liu, Yinuo and Zhou, Huichi and Zhang, Qihui and Wan, Yao and Zhou, Pan and Sun, Lichao},
title = {MLLM-as-a-Judge: assessing multimodal LLM-as-a-Judge with vision-language benchmark},
year = {2024},
publisher = {JMLR.org},
booktitle = {Proceedings of the 41st International Conference on Machine Learning},
articleno = {254},
numpages = {34},
location = {Vienna, Austria},
series = {ICML'24}
}

@inproceedings{yu2024fincon,
author = {Yu, Yangyang and Yao, Zhiyuan and Li, Haohang and Deng, Zhiyang and Jiang, Yuechen and Cao, Yupeng and Chen, Zhi and Suchow, Jordan W. and Cui, Zhenyu and Liu, Rong and Xu, Zhaozhuo and Zhang, Denghui and Subbalakshmi, Koduvayur and Xiong, Guojun and He, Yueru and Huang, Jimin and Li, Dong and Xie, Qianqian},
title = {FINCON: a synthesized LLM multi-agent system with conceptual verbal reinforcement for enhanced financial decision making},
year = {2024},
isbn = {9798331314385},
publisher = {Curran Associates Inc.},
address = {Red Hook, NY, USA},
booktitle = {Proceedings of the 38th International Conference on Neural Information Processing Systems},
articleno = {4354},
numpages = {36},
location = {Vancouver, BC, Canada},
series = {NIPS '24}
}

@inproceedings{cheong2024not,
author = {Cheong, Inyoung and Xia, King and Feng, K. J. Kevin and Chen, Quan Ze and Zhang, Amy X.},
title = {(A)I Am Not a Lawyer, But...: Engaging Legal Experts towards Responsible LLM Policies for Legal Advice},
year = {2024},
isbn = {9798400704505},
publisher = {Association for Computing Machinery},
address = {New York, NY, USA},
url = {https://doi.org/10.1145/3630106.3659048},
doi = {10.1145/3630106.3659048},
booktitle = {Proceedings of the 2024 ACM Conference on Fairness, Accountability, and Transparency},
pages = {2454–2469},
numpages = {16},
location = {Rio de Janeiro, Brazil},
series = {FAccT '24}
}

@inproceedings{cong2026estimatingllmgradingability,
    title = "Estimating {LLM} Grading Ability and Response Difficulty in Automatic Short Answer Grading via Item Response Theory",
    author = "Cong, Longwei  and
      Hahn, Sonja  and
      Gombert, Sebastian  and
      Camus, Leon  and
      Drachsler, Hendrik  and
      Kroehne, Ulf",
    editor = "Kochmar, Ekaterina  and
      Alhafni, Bashar  and
      Bann{\`o}, Stefano  and
      Bexte, Marie  and
      Burstein, Jill  and
      Horbach, Andrea  and
      Laarmann-Quante, Ronja  and
      Tack, Anais  and
      Yaneva, Victoria  and
      Yuan, Zheng",
    booktitle = "Proceedings of the 21st Workshop on Innovative Use of {NLP} for Building Educational Applications ({BEA} 2026)",
    month = jul,
    year = "2026",
    address = "San Diego, California, USA",
    publisher = "Association for Computational Linguistics",
    url = "https://aclanthology.org/2026.bea-1.19/",
    doi = "10.18653/v1/2026.bea-1.19",
    pages = "259--271",
    ISBN = "979-8-89176-409-5",
}

@article{naveed2025comprehensive,
author = {Naveed, Humza and Khan, Asad Ullah and Qiu, Shi and Saqib, Muhammad and Anwar, Saeed and Usman, Muhammad and Akhtar, Naveed and Barnes, Nick and Mian, Ajmal},
title = {A Comprehensive Overview of Large Language Models},
year = {2025},
issue_date = {October 2025},
publisher = {Association for Computing Machinery},
address = {New York, NY, USA},
volume = {16},
number = {5},
issn = {2157-6904},
url = {https://doi.org/10.1145/3744746},
doi = {10.1145/3744746},
journal = {ACM Trans. Intell. Syst. Technol.},
month = aug,
articleno = {106},
numpages = {72}
}

@inproceedings{yu-etal-2025-improve,
    title = "Improve {LLM}-as-a-Judge Ability as a General Ability",
    author = "Yu, Jiachen  and
      Sun, Shaoning  and
      Hu, Xiaohui  and
      Yan, Jiaxu  and
      Yu, Kaidong  and
      Li, Xuelong",
    editor = "Christodoulopoulos, Christos  and
      Chakraborty, Tanmoy  and
      Rose, Carolyn  and
      Peng, Violet",
    booktitle = "Proceedings of the 2025 Conference on Empirical Methods in Natural Language Processing",
    month = nov,
    year = "2025",
    address = "Suzhou, China",
    publisher = "Association for Computational Linguistics",
    url = "https://aclanthology.org/2025.emnlp-main.712/",
    doi = "10.18653/v1/2025.emnlp-main.712",
    pages = "14099--14115",
    ISBN = "979-8-89176-332-6"
}

@inproceedings{pan-etal-2024-human,
    title = "Human-Centered Design Recommendations for {LLM}-as-a-judge",
    author = "Pan, Qian  and
      Ashktorab, Zahra  and
      Desmond, Michael  and
      Santill{\'a}n Cooper, Mart{\'i}n  and
      Johnson, James  and
      Nair, Rahul  and
      Daly, Elizabeth  and
      Geyer, Werner",
    editor = "Soni, Nikita  and
      Flek, Lucie  and
      Sharma, Ashish  and
      Yang, Diyi  and
      Hooker, Sara  and
      Schwartz, H. Andrew",
    booktitle = "Proceedings of the 1st Human-Centered Large Language Modeling Workshop",
    month = aug,
    year = "2024",
    publisher = "ACL",
    url = "https://aclanthology.org/2024.hucllm-1.2/",
    doi = "10.18653/v1/2024.hucllm-1.2",
    pages = "16--29"
}

@misc{ye2026largelanguagemodelpsychometrics,
      title={Large Language Model Psychometrics: A Systematic Review of Evaluation, Validation, and Enhancement}, 
      author={Haoran Ye and Jing Jin and Yuhang Xie and Xin Zhang and Guojie Song},
      year={2026},
      eprint={2505.08245},
      archivePrefix={arXiv},
      primaryClass={cs.CL},
      url={https://arxiv.org/abs/2505.08245}, 
}

@inproceedings{kryscinski-etal-2020-evaluating,
    title = "Evaluating the Factual Consistency of Abstractive Text Summarization",
    author = "Kryscinski, Wojciech  and
      McCann, Bryan  and
      Xiong, Caiming  and
      Socher, Richard",
    editor = "Webber, Bonnie  and
      Cohn, Trevor  and
      He, Yulan  and
      Liu, Yang",
    booktitle = "Proceedings of the 2020 Conference on Empirical Methods in Natural Language Processing (EMNLP)",
    month = nov,
    year = "2020",
    address = "Online",
    publisher = "Association for Computational Linguistics",
    url = "https://aclanthology.org/2020.emnlp-main.750/",
    doi = "10.18653/v1/2020.emnlp-main.750",
    pages = "9332--9346"
}

@inproceedings{wang-etal-2020-asking,
    title = "Asking and Answering Questions to Evaluate the Factual Consistency of Summaries",
    author = "Wang, Alex  and
      Cho, Kyunghyun  and
      Lewis, Mike",
    editor = "Jurafsky, Dan  and
      Chai, Joyce  and
      Schluter, Natalie  and
      Tetreault, Joel",
    booktitle = "Proceedings of the 58th Annual Meeting of the Association for Computational Linguistics",
    month = jul,
    year = "2020",
    address = "Online",
    publisher = "Association for Computational Linguistics",
    url = "https://aclanthology.org/2020.acl-main.450/",
    doi = "10.18653/v1/2020.acl-main.450",
    pages = "5008--5020"
}

@inproceedings{zhao-etal-2023-discoscore,
    title = "{D}isco{S}core: Evaluating Text Generation with {BERT} and Discourse Coherence",
    author = "Zhao, Wei  and
      Strube, Michael  and
      Eger, Steffen",
    editor = "Vlachos, Andreas  and
      Augenstein, Isabelle",
    booktitle = "Proceedings of the 17th Conference of the European Chapter of the Association for Computational Linguistics",
    month = may,
    year = "2023",
    address = "Dubrovnik, Croatia",
    publisher = "Association for Computational Linguistics",
    url = "https://aclanthology.org/2023.eacl-main.278/",
    doi = "10.18653/v1/2023.eacl-main.278",
    pages = "3865--3883"
}

@inproceedings{barzilay-lapata-2005-modeling,
    title = "Modeling Local Coherence: An Entity-Based Approach",
    author = "Barzilay, Regina  and
      Lapata, Mirella",
    editor = "Knight, Kevin  and
      Ng, Hwee Tou  and
      Oflazer, Kemal",
    booktitle = "Proceedings of the 43rd Annual Meeting of the Association for Computational Linguistics ({ACL}{'}05)",
    month = jun,
    year = "2005",
    address = "Ann Arbor, Michigan",
    publisher = "Association for Computational Linguistics",
    url = "https://aclanthology.org/P05-1018/",
    doi = "10.3115/1219840.1219858",
    pages = "141--148"
}

@inproceedings{10.5555/3295222.3295230,
author = {Lundberg, Scott M. and Lee, Su-In},
title = {A unified approach to interpreting model predictions},
year = {2017},
isbn = {9781510860964},
publisher = {Curran Associates Inc.},
address = {Red Hook, NY, USA},
booktitle = {Proceedings of the 31st International Conference on Neural Information Processing Systems},
pages = {4768–4777},
numpages = {10},
location = {Long Beach, California, USA},
series = {NIPS'17}
}

@inproceedings{ye2025justice,
 author = {Ye, Jiayi and Wang, Yanbo and Huang, Yue and Chen, Dongping and Zhang, Qihui and Moniz, Nuno and Gao, Tian and Geyer, Werner and Huang, Chao and Chen, Pin-Yu and Chawla, Nitesh and Zhang, Xiangliang},
 booktitle = {International Conference on Learning Representations},
 editor = {Y. Yue and A. Garg and N. Peng and F. Sha and R. Yu},
 pages = {102351--102390},
 title = {Justice or Prejudice? Quantifying Biases in LLM-as-a-Judge},
 url = {https://proceedings.iclr.cc/paper_files/paper/2025/file/fdca08d371e4b6c031397909e20043bd-Paper-Conference.pdf},
 volume = {2025},
 year = {2025}
}

@inproceedings{shen-etal-2023-large,
    title = "Large Language Models are Not Yet Human-Level Evaluators for Abstractive Summarization",
    author = "Shen, Chenhui  and
      Cheng, Liying  and
      Nguyen, Xuan-Phi  and
      You, Yang  and
      Bing, Lidong",
    editor = "Bouamor, Houda  and
      Pino, Juan  and
      Bali, Kalika",
    booktitle = "Findings of the Association for Computational Linguistics: EMNLP 2023",
    month = dec,
    year = "2023",
    address = "Singapore",
    publisher = "Association for Computational Linguistics",
    url = "https://aclanthology.org/2023.findings-emnlp.278/",
    doi = "10.18653/v1/2023.findings-emnlp.278",
    pages = "4215--4233"
}

@misc{chen2023evaluatingfactualconsistencysummaries,
      title={Evaluating Factual Consistency of Summaries with Large Language Models}, 
      author={Shiqi Chen and Siyang Gao and Junxian He},
      year={2023},
      eprint={2305.14069},
      archivePrefix={arXiv},
      primaryClass={cs.CL},
      url={https://arxiv.org/abs/2305.14069}, 
}

@inproceedings{gabriel-etal-2021-go,
    title = "{GO} {FIGURE}: A Meta Evaluation of Factuality in Summarization",
    author = "Gabriel, Saadia  and
      Celikyilmaz, Asli  and
      Jha, Rahul  and
      Choi, Yejin  and
      Gao, Jianfeng",
    editor = "Zong, Chengqing  and
      Xia, Fei  and
      Li, Wenjie  and
      Navigli, Roberto",
    booktitle = "Findings of the Association for Computational Linguistics: ACL-IJCNLP 2021",
    month = aug,
    year = "2021",
    address = "Online",
    publisher = "Association for Computational Linguistics",
    url = "https://aclanthology.org/2021.findings-acl.42/",
    doi = "10.18653/v1/2021.findings-acl.42",
    pages = "478--487"
}

@misc{Kamath2025Gemma3T,
      title={Gemma 3 Technical Report}, 
      author={Gemma Team and Aishwarya Kamath and Johan Ferret and Shreya Pathak and Nino Vieillard and Ramona Merhej and Sarah Perrin and Tatiana Matejovicova and Alexandre Ramé and Morgane Rivière and Louis Rouillard and Thomas Mesnard and Geoffrey Cideron and Jean-bastien Grill and Sabela Ramos and Edouard Yvinec and Michelle Casbon and Etienne Pot and Ivo Penchev and Gaël Liu and Francesco Visin and Kathleen Kenealy and Lucas Beyer and Xiaohai Zhai and Anton Tsitsulin and Robert Busa-Fekete and Alex Feng and Noveen Sachdeva and Benjamin Coleman and Yi Gao and Basil Mustafa and Iain Barr and Emilio Parisotto and David Tian and Matan Eyal and Colin Cherry and Jan-Thorsten Peter and Danila Sinopalnikov and Surya Bhupatiraju and Rishabh Agarwal and Mehran Kazemi and Dan Malkin and Ravin Kumar and David Vilar and Idan Brusilovsky and Jiaming Luo and Andreas Steiner and Abe Friesen and Abhanshu Sharma and Abheesht Sharma and Adi Mayrav Gilady and Adrian Goedeckemeyer and Alaa Saade and Alex Feng and Alexander Kolesnikov and Alexei Bendebury and Alvin Abdagic and Amit Vadi and András György and André Susano Pinto and Anil Das and Ankur Bapna and Antoine Miech and Antoine Yang and Antonia Paterson and Ashish Shenoy and Ayan Chakrabarti and Bilal Piot and Bo Wu and Bobak Shahriari and Bryce Petrini and Charlie Chen and Charline Le Lan and Christopher A. Choquette-Choo and CJ Carey and Cormac Brick and Daniel Deutsch and Danielle Eisenbud and Dee Cattle and Derek Cheng and Dimitris Paparas and Divyashree Shivakumar Sreepathihalli and Doug Reid and Dustin Tran and Dustin Zelle and Eric Noland and Erwin Huizenga and Eugene Kharitonov and Frederick Liu and Gagik Amirkhanyan and Glenn Cameron and Hadi Hashemi and Hanna Klimczak-Plucińska and Harman Singh and Harsh Mehta and Harshal Tushar Lehri and Hussein Hazimeh and Ian Ballantyne and Idan Szpektor and Ivan Nardini and Jean Pouget-Abadie and Jetha Chan and Joe Stanton and John Wieting and Jonathan Lai and Jordi Orbay and Joseph Fernandez and Josh Newlan and Ju-yeong Ji and Jyotinder Singh and Kat Black and Kathy Yu and Kevin Hui and Kiran Vodrahalli and Klaus Greff and Linhai Qiu and Marcella Valentine and Marina Coelho and Marvin Ritter and Matt Hoffman and Matthew Watson and Mayank Chaturvedi and Michael Moynihan and Min Ma and Nabila Babar and Natasha Noy and Nathan Byrd and Nick Roy and Nikola Momchev and Nilay Chauhan and Noveen Sachdeva and Oskar Bunyan and Pankil Botarda and Paul Caron and Paul Kishan Rubenstein and Phil Culliton and Philipp Schmid and Pier Giuseppe Sessa and Pingmei Xu and Piotr Stanczyk and Pouya Tafti and Rakesh Shivanna and Renjie Wu and Renke Pan and Reza Rokni and Rob Willoughby and Rohith Vallu and Ryan Mullins and Sammy Jerome and Sara Smoot and Sertan Girgin and Shariq Iqbal and Shashir Reddy and Shruti Sheth and Siim Põder and Sijal Bhatnagar and Sindhu Raghuram Panyam and Sivan Eiger and Susan Zhang and Tianqi Liu and Trevor Yacovone and Tyler Liechty and Uday Kalra and Utku Evci and Vedant Misra and Vincent Roseberry and Vlad Feinberg and Vlad Kolesnikov and Woohyun Han and Woosuk Kwon and Xi Chen and Yinlam Chow and Yuvein Zhu and Zichuan Wei and Zoltan Egyed and Victor Cotruta and Minh Giang and Phoebe Kirk and Anand Rao and Kat Black and Nabila Babar and Jessica Lo and Erica Moreira and Luiz Gustavo Martins and Omar Sanseviero and Lucas Gonzalez and Zach Gleicher and Tris Warkentin and Vahab Mirrokni and Evan Senter and Eli Collins and Joelle Barral and Zoubin Ghahramani and Raia Hadsell and Yossi Matias and D. Sculley and Slav Petrov and Noah Fiedel and Noam Shazeer and Oriol Vinyals and Jeff Dean and Demis Hassabis and Koray Kavukcuoglu and Clement Farabet and Elena Buchatskaya and Jean-Baptiste Alayrac and Rohan Anil and Dmitry and Lepikhin and Sebastian Borgeaud and Olivier Bachem and Armand Joulin and Alek Andreev and Cassidy Hardin and Robert Dadashi and Léonard Hussenot},
      year={2025},
      eprint={2503.19786},
      archivePrefix={arXiv},
      primaryClass={cs.CL},
      url={https://arxiv.org/abs/2503.19786}, 
}

@misc{qwen3.5,
    title  = {{Qwen3.5}: Towards Native Multimodal Agents},
    author = {{Qwen Team}},
    month  = {February},
    year   = {2026},
    url    = {https://qwen.ai/blog?id=qwen3.5}
}

@misc{abdin2024phi,
      title={Phi-4 Technical Report}, 
      author={Marah Abdin and Jyoti Aneja and Harkirat Behl and Sébastien Bubeck and Ronen Eldan and Suriya Gunasekar and Michael Harrison and Russell J. Hewett and Mojan Javaheripi and Piero Kauffmann and James R. Lee and Yin Tat Lee and Yuanzhi Li and Weishung Liu and Caio C. T. Mendes and Anh Nguyen and Eric Price and Gustavo de Rosa and Olli Saarikivi and Adil Salim and Shital Shah and Xin Wang and Rachel Ward and Yue Wu and Dingli Yu and Cyril Zhang and Yi Zhang},
      year={2024},
      eprint={2412.08905},
      archivePrefix={arXiv},
      primaryClass={cs.CL},
      url={https://arxiv.org/abs/2412.08905}, 
}

@misc{grattafiori2024llama,
      title={The Llama 3 Herd of Models}, 
      author={Aaron Grattafiori and Abhimanyu Dubey and Abhinav Jauhri and Abhinav Pandey and Abhishek Kadian and Ahmad Al-Dahle and Aiesha Letman and Akhil Mathur and Alan Schelten and Alex Vaughan and Amy Yang and Angela Fan and Anirudh Goyal and Anthony Hartshorn and Aobo Yang and Archi Mitra and Archie Sravankumar and Artem Korenev and Arthur Hinsvark and Arun Rao and Aston Zhang and Aurelien Rodriguez and Austen Gregerson and Ava Spataru and Baptiste Roziere and Bethany Biron and Binh Tang and Bobbie Chern and Charlotte Caucheteux and Chaya Nayak and Chloe Bi and Chris Marra and Chris McConnell and Christian Keller and Christophe Touret and Chunyang Wu and Corinne Wong and Cristian Canton Ferrer and Cyrus Nikolaidis and Damien Allonsius and Daniel Song and Danielle Pintz and Danny Livshits and Danny Wyatt and David Esiobu and Dhruv Choudhary and Dhruv Mahajan and Diego Garcia-Olano and Diego Perino and Dieuwke Hupkes and Egor Lakomkin and Ehab AlBadawy and Elina Lobanova and Emily Dinan and Eric Michael Smith and Filip Radenovic and Francisco Guzmán and Frank Zhang and Gabriel Synnaeve and Gabrielle Lee and Georgia Lewis Anderson and Govind Thattai and Graeme Nail and Gregoire Mialon and Guan Pang and Guillem Cucurell and Hailey Nguyen and Hannah Korevaar and Hu Xu and Hugo Touvron and Iliyan Zarov and Imanol Arrieta Ibarra and Isabel Kloumann and Ishan Misra and Ivan Evtimov and Jack Zhang and Jade Copet and Jaewon Lee and Jan Geffert and Jana Vranes and Jason Park and Jay Mahadeokar and Jeet Shah and Jelmer van der Linde and Jennifer Billock and Jenny Hong and Jenya Lee and Jeremy Fu and Jianfeng Chi and Jianyu Huang and Jiawen Liu and Jie Wang and Jiecao Yu and Joanna Bitton and Joe Spisak and Jongsoo Park and Joseph Rocca and Joshua Johnstun and Joshua Saxe and Junteng Jia and Kalyan Vasuden Alwala and Karthik Prasad and Kartikeya Upasani and Kate Plawiak and Ke Li and Kenneth Heafield and Kevin Stone and Khalid El-Arini and Krithika Iyer and Kshitiz Malik and Kuenley Chiu and Kunal Bhalla and Kushal Lakhotia and Lauren Rantala-Yeary and Laurens van der Maaten and Lawrence Chen and Liang Tan and Liz Jenkins and Louis Martin and Lovish Madaan and Lubo Malo and Lukas Blecher and Lukas Landzaat and Luke de Oliveira and Madeline Muzzi and Mahesh Pasupuleti and Mannat Singh and Manohar Paluri and Marcin Kardas and Maria Tsimpoukelli and Mathew Oldham and Mathieu Rita and Maya Pavlova and Melanie Kambadur and Mike Lewis and Min Si and Mitesh Kumar Singh and Mona Hassan and Naman Goyal and Narjes Torabi and Nikolay Bashlykov and Nikolay Bogoychev and Niladri Chatterji and Ning Zhang and Olivier Duchenne and Onur Çelebi and Patrick Alrassy and Pengchuan Zhang and Pengwei Li and Petar Vasic and Peter Weng and Prajjwal Bhargava and Pratik Dubal and Praveen Krishnan and Punit Singh Koura and Puxin Xu and Qing He and Qingxiao Dong and Ragavan Srinivasan and Raj Ganapathy and Ramon Calderer and Ricardo Silveira Cabral and Robert Stojnic and Roberta Raileanu and Rohan Maheswari and Rohit Girdhar and Rohit Patel and Romain Sauvestre and Ronnie Polidoro and Roshan Sumbaly and Ross Taylor and Ruan Silva and Rui Hou and Rui Wang and Saghar Hosseini and Sahana Chennabasappa and Sanjay Singh and Sean Bell and Seohyun Sonia Kim and Sergey Edunov and Shaoliang Nie and Sharan Narang and Sharath Raparthy and Sheng Shen and Shengye Wan and Shruti Bhosale and Shun Zhang and Simon Vandenhende and Soumya Batra and Spencer Whitman and Sten Sootla and Stephane Collot and Suchin Gururangan and Sydney Borodinsky and Tamar Herman and Tara Fowler and Tarek Sheasha and Thomas Georgiou and Thomas Scialom and Tobias Speckbacher and Todor Mihaylov and Tong Xiao and Ujjwal Karn and Vedanuj Goswami and Vibhor Gupta and Vignesh Ramanathan and Viktor Kerkez and Vincent Gonguet and Virginie Do and Vish Vogeti and Vítor Albiero and Vladan Petrovic and Weiwei Chu and Wenhan Xiong and Wenyin Fu and Whitney Meers and Xavier Martinet and Xiaodong Wang and Xiaofang Wang and Xiaoqing Ellen Tan and Xide Xia and Xinfeng Xie and Xuchao Jia and Xuewei Wang and Yaelle Goldschlag and Yashesh Gaur and Yasmine Babaei and Yi Wen and Yiwen Song and Yuchen Zhang and Yue Li and Yuning Mao and Zacharie Delpierre Coudert and Zheng Yan and Zhengxing Chen and Zoe Papakipos and Aaditya Singh and Aayushi Srivastava and Abha Jain and Adam Kelsey and Adam Shajnfeld and Adithya Gangidi and Adolfo Victoria and Ahuva Goldstand and Ajay Menon and Ajay Sharma and Alex Boesenberg and Alexei Baevski and Allie Feinstein and Amanda Kallet and Amit Sangani and Amos Teo and Anam Yunus and Andrei Lupu and Andres Alvarado and Andrew Caples and Andrew Gu and Andrew Ho and Andrew Poulton and Andrew Ryan and Ankit Ramchandani and Annie Dong and Annie Franco and Anuj Goyal and Aparajita Saraf and Arkabandhu Chowdhury and Ashley Gabriel and Ashwin Bharambe and Assaf Eisenman and Azadeh Yazdan and Beau James and Ben Maurer and Benjamin Leonhardi and Bernie Huang and Beth Loyd and Beto De Paola and Bhargavi Paranjape and Bing Liu and Bo Wu and Boyu Ni and Braden Hancock and Bram Wasti and Brandon Spence and Brani Stojkovic and Brian Gamido and Britt Montalvo and Carl Parker and Carly Burton and Catalina Mejia and Ce Liu and Changhan Wang and Changkyu Kim and Chao Zhou and Chester Hu and Ching-Hsiang Chu and Chris Cai and Chris Tindal and Christoph Feichtenhofer and Cynthia Gao and Damon Civin and Dana Beaty and Daniel Kreymer and Daniel Li and David Adkins and David Xu and Davide Testuggine and Delia David and Devi Parikh and Diana Liskovich and Didem Foss and Dingkang Wang and Duc Le and Dustin Holland and Edward Dowling and Eissa Jamil and Elaine Montgomery and Eleonora Presani and Emily Hahn and Emily Wood and Eric-Tuan Le and Erik Brinkman and Esteban Arcaute and Evan Dunbar and Evan Smothers and Fei Sun and Felix Kreuk and Feng Tian and Filippos Kokkinos and Firat Ozgenel and Francesco Caggioni and Frank Kanayet and Frank Seide and Gabriela Medina Florez and Gabriella Schwarz and Gada Badeer and Georgia Swee and Gil Halpern and Grant Herman and Grigory Sizov and Guangyi and Zhang and Guna Lakshminarayanan and Hakan Inan and Hamid Shojanazeri and Han Zou and Hannah Wang and Hanwen Zha and Haroun Habeeb and Harrison Rudolph and Helen Suk and Henry Aspegren and Hunter Goldman and Hongyuan Zhan and Ibrahim Damlaj and Igor Molybog and Igor Tufanov and Ilias Leontiadis and Irina-Elena Veliche and Itai Gat and Jake Weissman and James Geboski and James Kohli and Janice Lam and Japhet Asher and Jean-Baptiste Gaya and Jeff Marcus and Jeff Tang and Jennifer Chan and Jenny Zhen and Jeremy Reizenstein and Jeremy Teboul and Jessica Zhong and Jian Jin and Jingyi Yang and Joe Cummings and Jon Carvill and Jon Shepard and Jonathan McPhie and Jonathan Torres and Josh Ginsburg and Junjie Wang and Kai Wu and Kam Hou U and Karan Saxena and Kartikay Khandelwal and Katayoun Zand and Kathy Matosich and Kaushik Veeraraghavan and Kelly Michelena and Keqian Li and Kiran Jagadeesh and Kun Huang and Kunal Chawla and Kyle Huang and Lailin Chen and Lakshya Garg and Lavender A and Leandro Silva and Lee Bell and Lei Zhang and Liangpeng Guo and Licheng Yu and Liron Moshkovich and Luca Wehrstedt and Madian Khabsa and Manav Avalani and Manish Bhatt and Martynas Mankus and Matan Hasson and Matthew Lennie and Matthias Reso and Maxim Groshev and Maxim Naumov and Maya Lathi and Meghan Keneally and Miao Liu and Michael L. Seltzer and Michal Valko and Michelle Restrepo and Mihir Patel and Mik Vyatskov and Mikayel Samvelyan and Mike Clark and Mike Macey and Mike Wang and Miquel Jubert Hermoso and Mo Metanat and Mohammad Rastegari and Munish Bansal and Nandhini Santhanam and Natascha Parks and Natasha White and Navyata Bawa and Nayan Singhal and Nick Egebo and Nicolas Usunier and Nikhil Mehta and Nikolay Pavlovich Laptev and Ning Dong and Norman Cheng and Oleg Chernoguz and Olivia Hart and Omkar Salpekar and Ozlem Kalinli and Parkin Kent and Parth Parekh and Paul Saab and Pavan Balaji and Pedro Rittner and Philip Bontrager and Pierre Roux and Piotr Dollar and Polina Zvyagina and Prashant Ratanchandani and Pritish Yuvraj and Qian Liang and Rachad Alao and Rachel Rodriguez and Rafi Ayub and Raghotham Murthy and Raghu Nayani and Rahul Mitra and Rangaprabhu Parthasarathy and Raymond Li and Rebekkah Hogan and Robin Battey and Rocky Wang and Russ Howes and Ruty Rinott and Sachin Mehta and Sachin Siby and Sai Jayesh Bondu and Samyak Datta and Sara Chugh and Sara Hunt and Sargun Dhillon and Sasha Sidorov and Satadru Pan and Saurabh Mahajan and Saurabh Verma and Seiji Yamamoto and Sharadh Ramaswamy and Shaun Lindsay and Shaun Lindsay and Sheng Feng and Shenghao Lin and Shengxin Cindy Zha and Shishir Patil and Shiva Shankar and Shuqiang Zhang and Shuqiang Zhang and Sinong Wang and Sneha Agarwal and Soji Sajuyigbe and Soumith Chintala and Stephanie Max and Stephen Chen and Steve Kehoe and Steve Satterfield and Sudarshan Govindaprasad and Sumit Gupta and Summer Deng and Sungmin Cho and Sunny Virk and Suraj Subramanian and Sy Choudhury and Sydney Goldman and Tal Remez and Tamar Glaser and Tamara Best and Thilo Koehler and Thomas Robinson and Tianhe Li and Tianjun Zhang and Tim Matthews and Timothy Chou and Tzook Shaked and Varun Vontimitta and Victoria Ajayi and Victoria Montanez and Vijai Mohan and Vinay Satish Kumar and Vishal Mangla and Vlad Ionescu and Vlad Poenaru and Vlad Tiberiu Mihailescu and Vladimir Ivanov and Wei Li and Wenchen Wang and Wenwen Jiang and Wes Bouaziz and Will Constable and Xiaocheng Tang and Xiaojian Wu and Xiaolan Wang and Xilun Wu and Xinbo Gao and Yaniv Kleinman and Yanjun Chen and Ye Hu and Ye Jia and Ye Qi and Yenda Li and Yilin Zhang and Ying Zhang and Yossi Adi and Youngjin Nam and Yu and Wang and Yu Zhao and Yuchen Hao and Yundi Qian and Yunlu Li and Yuzi He and Zach Rait and Zachary DeVito and Zef Rosnbrick and Zhaoduo Wen and Zhenyu Yang and Zhiwei Zhao and Zhiyu Ma},
      year={2024},
      eprint={2407.21783},
      archivePrefix={arXiv},
      primaryClass={cs.AI},
      url={https://arxiv.org/abs/2407.21783}, 
}

@misc{liu2026ministral,
      title={Ministral 3}, 
      author={Alexander H. Liu and Kartik Khandelwal and Sandeep Subramanian and Victor Jouault and Abhinav Rastogi and Adrien Sadé and Alan Jeffares and Albert Jiang and Alexandre Cahill and Alexandre Gavaudan and Alexandre Sablayrolles and Amélie Héliou and Amos You and Andy Ehrenberg and Andy Lo and Anton Eliseev and Antonia Calvi and Avinash Sooriyarachchi and Baptiste Bout and Baptiste Rozière and Baudouin De Monicault and Clémence Lanfranchi and Corentin Barreau and Cyprien Courtot and Daniele Grattarola and Darius Dabert and Diego de las Casas and Elliot Chane-Sane and Faruk Ahmed and Gabrielle Berrada and Gaëtan Ecrepont and Gauthier Guinet and Georgii Novikov and Guillaume Kunsch and Guillaume Lample and Guillaume Martin and Gunshi Gupta and Jan Ludziejewski and Jason Rute and Joachim Studnia and Jonas Amar and Joséphine Delas and Josselin Somerville Roberts and Karmesh Yadav and Khyathi Chandu and Kush Jain and Laurence Aitchison and Laurent Fainsin and Léonard Blier and Lingxiao Zhao and Louis Martin and Lucile Saulnier and Luyu Gao and Maarten Buyl and Margaret Jennings and Marie Pellat and Mark Prins and Mathieu Poirée and Mathilde Guillaumin and Matthieu Dinot and Matthieu Futeral and Maxime Darrin and Maximilian Augustin and Mia Chiquier and Michel Schimpf and Nathan Grinsztajn and Neha Gupta and Nikhil Raghuraman and Olivier Bousquet and Olivier Duchenne and Patricia Wang and Patrick von Platen and Paul Jacob and Paul Wambergue and Paula Kurylowicz and Pavankumar Reddy Muddireddy and Philomène Chagniot and Pierre Stock and Pravesh Agrawal and Quentin Torroba and Romain Sauvestre and Roman Soletskyi and Rupert Menneer and Sagar Vaze and Samuel Barry and Sanchit Gandhi and Siddhant Waghjale and Siddharth Gandhi and Soham Ghosh and Srijan Mishra and Sumukh Aithal and Szymon Antoniak and Teven Le Scao and Théo Cachet and Theo Simon Sorg and Thibaut Lavril and Thiziri Nait Saada and Thomas Chabal and Thomas Foubert and Thomas Robert and Thomas Wang and Tim Lawson and Tom Bewley and Tom Bewley and Tom Edwards and Umar Jamil and Umberto Tomasini and Valeriia Nemychnikova and Van Phung and Vincent Maladière and Virgile Richard and Wassim Bouaziz and Wen-Ding Li and William Marshall and Xinghui Li and Xinyu Yang and Yassine El Ouahidi and Yihan Wang and Yunhao Tang and Zaccharie Ramzi},
      year={2026},
      eprint={2601.08584},
      archivePrefix={arXiv},
      primaryClass={cs.CL},
      url={https://arxiv.org/abs/2601.08584}, 
}

@inproceedings{ICLR2024_fc8781fb,
 author = {Wang, Guan and Cheng, Sijie and Zhan, Xianyuan and Li, Xiangang and Song, Sen and Liu, Yang},
 booktitle = {International Conference on Learning Representations},
 editor = {B. Kim and Y. Yue and S. Chaudhuri and K. Fragkiadaki and M. Khan and Y. Sun},
 pages = {57021--57040},
 title = {OpenChat: Advancing Open-source Language Models with Mixed-Quality Data},
 url = {https://proceedings.iclr.cc/paper_files/paper/2024/file/fc8781fb328fb1fd069584a4519a2709-Paper-Conference.pdf},
 volume = {2024},
 year = {2024}
}

@misc{abdin2024phi3technicalreporthighly,
      title={Phi-3 Technical Report: A Highly Capable Language Model Locally on Your Phone}, 
      author={Marah Abdin and Jyoti Aneja and Hany Awadalla and Ahmed Awadallah and Ammar Ahmad Awan and Nguyen Bach and Amit Bahree and Arash Bakhtiari and Jianmin Bao and Harkirat Behl and Alon Benhaim and Misha Bilenko and Johan Bjorck and Sébastien Bubeck and Martin Cai and Qin Cai and Vishrav Chaudhary and Dong Chen and Dongdong Chen and Weizhu Chen and Yen-Chun Chen and Yi-Ling Chen and Hao Cheng and Parul Chopra and Xiyang Dai and Matthew Dixon and Ronen Eldan and Victor Fragoso and Jianfeng Gao and Mei Gao and Min Gao and Amit Garg and Allie Del Giorno and Abhishek Goswami and Suriya Gunasekar and Emman Haider and Junheng Hao and Russell J. Hewett and Wenxiang Hu and Jamie Huynh and Dan Iter and Sam Ade Jacobs and Mojan Javaheripi and Xin Jin and Nikos Karampatziakis and Piero Kauffmann and Mahoud Khademi and Dongwoo Kim and Young Jin Kim and Lev Kurilenko and James R. Lee and Yin Tat Lee and Yuanzhi Li and Yunsheng Li and Chen Liang and Lars Liden and Xihui Lin and Zeqi Lin and Ce Liu and Liyuan Liu and Mengchen Liu and Weishung Liu and Xiaodong Liu and Chong Luo and Piyush Madan and Ali Mahmoudzadeh and David Majercak and Matt Mazzola and Caio César Teodoro Mendes and Arindam Mitra and Hardik Modi and Anh Nguyen and Brandon Norick and Barun Patra and Daniel Perez-Becker and Thomas Portet and Reid Pryzant and Heyang Qin and Marko Radmilac and Liliang Ren and Gustavo de Rosa and Corby Rosset and Sambudha Roy and Olatunji Ruwase and Olli Saarikivi and Amin Saied and Adil Salim and Michael Santacroce and Shital Shah and Ning Shang and Hiteshi Sharma and Yelong Shen and Swadheen Shukla and Xia Song and Masahiro Tanaka and Andrea Tupini and Praneetha Vaddamanu and Chunyu Wang and Guanhua Wang and Lijuan Wang and Shuohang Wang and Xin Wang and Yu Wang and Rachel Ward and Wen Wen and Philipp Witte and Haiping Wu and Xiaoxia Wu and Michael Wyatt and Bin Xiao and Can Xu and Jiahang Xu and Weijian Xu and Jilong Xue and Sonali Yadav and Fan Yang and Jianwei Yang and Yifan Yang and Ziyi Yang and Donghan Yu and Lu Yuan and Chenruidong Zhang and Cyril Zhang and Jianwen Zhang and Li Lyna Zhang and Yi Zhang and Yue Zhang and Yunan Zhang and Xiren Zhou},
      year={2024},
      eprint={2404.14219},
      archivePrefix={arXiv},
      primaryClass={cs.CL},
      url={https://arxiv.org/abs/2404.14219}, 
}
